\documentclass[10pt,twocolumn,letterpaper]{article}

\usepackage[pagenumbers]{cvpr} %

\usepackage{titletoc} %
\usepackage{xspace}
\usepackage{multirow} %
\usepackage{colortbl} %
\usepackage{soul} %
\usepackage{fancyvrb} %
\usepackage{listings} %
\usepackage[accsupp]{axessibility}  %

\newcommand{\z}{\textcolor{white}{0}} %

\usepackage{pifont}
\newcommand{\cmark}{\ding{51}} %
\newcommand{\xmark}{\ding{55}} %

\definecolor{lightgray}{gray}{0.9}
\definecolor{lightgreen}{rgb}{0.9, 0.9, 1}
\definecolor{lightred}{rgb}{1, 0.8, 0.8}
\definecolor{darkgreen}{rgb}{0, 0.5, 0}
\definecolor{backcolour}{rgb}{0.95,0.95,0.92}
\definecolor{cvprblue}{rgb}{0.21,0.49,0.74}
\newcommand{\new}[1]{{\color[rgb]{0.0,0.0,0.0}#1}} %
\newcommand{\method}{Chapter-Llama\xspace}

\newcommand{\ourtitle}{\method: Efficient 
Chaptering in Hour-Long Videos with LLMs} %

\newcommand{\gray}[1]{{\color{gray}#1}} %

\usepackage[subtle]{savetrees} %

\newcommand{\appendixref}[2]{%
  \if\sepappendix1%
    #1%
  \else%
    #2%
  \fi%
}

\def\sepappendix{0}

\definecolor{cvprblue}{rgb}{0.21,0.49,0.74}
\usepackage[pagebackref,breaklinks,colorlinks,allcolors=cvprblue]{hyperref}

\def\confName{CVPR}
\def\confYear{2025}

\title{\ourtitle}

\author{Lucas Ventura\textsuperscript{1,2}
\hspace{1em}
Antoine Yang\textsuperscript{3}
\hspace{1em}
Cordelia Schmid\textsuperscript{2}
\hspace{1em}
G\"ul Varol\textsuperscript{1}
\vspace{0.5mm}
\and
{\small\textsuperscript{1}LIGM, \'{E}cole des Ponts, IP Paris, Univ Gustave Eiffel, CNRS}
\\
{\small\textsuperscript{2}Inria, \'{E}cole normale sup\'{e}rieure, CNRS, PSL Research University}
\hspace{2mm}
{\small\textsuperscript{3}Google DeepMind}
\\
{\tt\small \url{https://imagine.enpc.fr/~lucas.ventura/chapter-llama/}}
}

\begin{document}
    \maketitle
    \begin{abstract}
We address the task of video chaptering, i.e.,
partitioning a long video timeline %
into semantic units and generating corresponding chapter titles.
While relatively underexplored, automatic chaptering
has the potential to enable efficient navigation and content
retrieval in long-form videos.
In this paper, we 
achieve 
strong
chaptering performance on hour-long videos
by efficiently addressing the problem in the text domain with our `\method' framework.
Specifically,
we leverage a pre-trained large language model (LLM) with large context window,
and feed as input (i)~speech transcripts 
and (ii)~captions describing video frames, along with their respective timestamps.
Given the inefficiency of exhaustively captioning all frames, we propose a
lightweight speech-guided frame
selection strategy
based on speech transcript content,
and experimentally demonstrate remarkable advantages. %
We train the LLM to output timestamps for the chapter boundaries, as well as free-form chapter titles.
This simple yet powerful approach scales to processing 
one-hour long videos in a single forward pass.
Our results 
demonstrate substantial
improvements %
(e.g., 45.3 vs 26.7 F1 score)
over the state of the art on the recent VidChapters-7M benchmark.
To promote further research, we release our code and models
\new{at our project page.}
\end{abstract}

    \section{Introduction}
\label{sec:intro}

According to a study by \cite{li2021}, the video durations uploaded to
the popular online video sharing platform YouTube %
have increased steadily over the years. Videos have become longer
since the first video upload in 2005 \cite{li2005,Cheng2008}. In 2020,
25\% of videos were estimated to be longer than 15~minutes,
5\% more than 3~hours \cite{li2021}.
Long-form videos
such as news, sports, educational, and vlog streams can
often span extensive durations and cover multiple topics \cite{violot2024shorts}.
Finding specific content within increased video duration and volume
makes efficient content navigation more important than ever.

However, much of the traditional video analysis research has focused
on processing \textit{short} videos of a few seconds \cite{bahdanau2014neural, rohrbach2013translating, sutskever2014sequence, pan2017video, seo2022end, pei2019memory, hori2017attention, yang2023vid2seq, luo2020univl, sun2019videobert, wang2022omnivl, chen2023valor}.
At the same time, the definition of \textit{long} videos has changed within the past decade.
Early works claimed processing 100 frames (i.e., a few seconds) to be long \cite{varol18_ltc,ng2015shortsnippetsdeepnetworks}
as opposed to ingesting up to 16 frames  \cite{simonyan2014,tran15c3d}.
With the introduction of datasets containing 1-5 minute videos \cite{sigurdsson16_charades,zhou18towards,krishna2017dense, huang2020multimodal,mangalam2023egoschema,Ego4D2022CVPR}, several minutes were considered \textit{very}~long.
Studying \mbox{\textit{hour-long}} videos has only recently seen an interest
in the context of movie description \cite{han23_autoad}, video captioning \cite{islam2024videoReCap},
or grounding \cite{soldan2022mad,hannan2023rgnet}. %
Very recently, the work of \cite{yang2023vidchapters} collected the VidChapters-7M dataset
with videos spanning from minutes to hours, along with their user-defined video \textit{chapters},
and proposed the video chapter generation task,
automatically dividing a video into thematic sections (i.e., chapters) with descriptive concise chapter titles.
Video chaptering, if achieved successfully, can offer a compelling solution
to long content indexing, bypassing the current need for time-consuming manual annotation by video owners
\cite{yang2023vidchapters}.

\begin{figure}
	\centering
	\includegraphics[width=.99\linewidth]{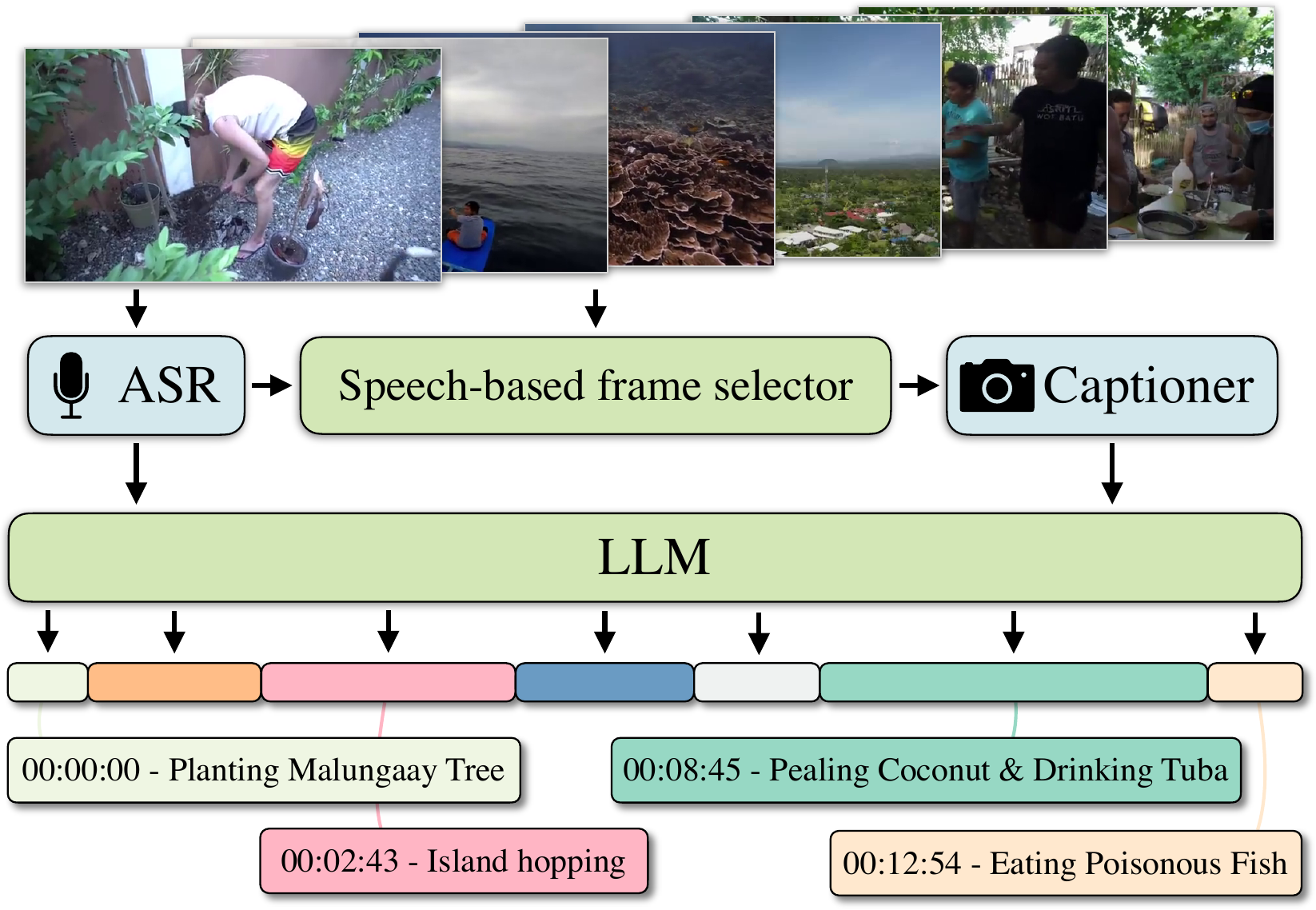}
        \vspace{-0.2cm}
	\caption{
		\textbf{Chapter-Llama:} 
		Our method generates automatic video chapters for hour-long videos by training a large language model (LLM) to predict chapter boundaries and titles.
        The LLM processes transcribed speech (ASR) and descriptive captions of key frames,
        which are %
        sampled based on ASR content.
        This
        \new{text-based}
        approach,
        \new{equipped with speech-based frame selection,}
        enables efficient processing of long-form content.
	}
\vspace{-0.2cm}
	\label{fig:teaser}
\end{figure}

In this paper, we address the challenge of automatic video chaptering 
with a simple yet effective framework designed to handle hour-long videos. %
Existing work for chaptering \cite{yang2023vidchapters} relies on a
dense video captioning model Vid2Seq~\cite{yang2023vid2seq}, 
which combines multimodal inputs 
from video frames and ASR-based speech transcriptions.
However, Vid2Seq operates on a fixed number of equally sampled frames (i.e., 100 frames),
potentially missing important visual information.
Furthermore, their approach based on
transformer architecture uses %
video frame features directly,
which requires learning a mapping from the visual modality to the textual modality.
In contrast, our method is designed to address these limitations
by (i)~dynamically sampling keyframes from the video based on the speech content,
and (ii)~designing a purely text-based model leveraging image captioning to convert RGB frames into text.

Our approach leverages a pretrained LLM,
which we finetune specifically for the video chaptering task
to predict jointly the chapter boundary timestamps and chapter titles,
both in text form.
The appeal of our model lies in processing only textual data as input, allowing us effectively leverage the long-context understanding capabilities of the LLM to scale to long videos.
In particular, we incorporate speech transcriptions from automatic speech recognition (ASR)
and automatic frame captions.
Captioning has been used for video understanding as an intermediate %
representation in recent works, but in the context of retrieval or question answering (QA)
for shorter videos (maximum 3~minutes) \cite{min2024morevqa,zeng2022socratic,ventura24multicaps,zhang2024simplellm}. %
In longer videos, since captioning every frame is computationally prohibitive,
we employ a speech-based frame selection strategy
that 
scales efficiently while preserving important content.
Similar in spirit to \cite{korbar2019scsampler}, we primarily use audio to determine keyframes,
specifically bootstrapping with an LLM trained only with the speech inputs.
However, even when transforming a video into text, 
LLMs have a limited context window, allowing a maximum number of tokens as input
in a single forward pass.
To mitigate context window limitations for very long video inputs, 
we simply perform an iterative prediction, sequentially processing the video,
where each iteration typically operates on a window length of about an hour duration.
We evaluate our approach on `short' (0-15 min), `medium' (15-30 min), and `long' (30-60 min) videos 
from the VidChapters-7M dataset \cite{yang2023vidchapters},
demonstrating significant improvements over the state of the art
across multiple metrics,
including temporal boundary accuracy and semantic relevance of chapter titles. %
Our experiments show that finetuning the LLM, our
speech-based frame selection strategy, and the integration of
modalities
from both speech and captions are crucial for achieving 
high-quality video chaptering results.

Our contributions are the following:
(i)~We introduce \method: our framework
leverages a pretrained LLM and finetunes for the underexplored task of video chaptering
by transforming the video input into \textit{text form through ASR and captioning}.
(ii)~We scale efficiently to hour-long videos by incorporating a \textit{speech-based frame sampling} strategy, captioning only a subset of the video frames.
(iii)~Our simple and effective approach \textit{
outperforms the state of the art} on the recent VidChapters-7M benchmark by a large margin (e.g., 45.3 vs 26.7 F1 score).
These results are complemented by a comprehensive set of experiments analyzing our components.

    \section{Related Work}
\label{sec:related}

We provide an overview of video tasks related to video chaptering,
such as temporal segmentation and captioning,
along with a discussion on works focusing
on long-form %
and LLM-based video understanding.

\vspace{-0.2cm}
\paragraph{Temporal video segmentation.} %
While video chaptering is a new task \cite{yang2023vidchapters},
there is a rich literature on methods focused on temporally segmenting a video in various forms. %
One task is \textit{shot detection}~\cite{rasheed2003scene, rui1998exploring, sidiropoulos2011temporal}, where
any visual changes (e.g., shifting between two cameras) would require a temporal boundary, not necessarily
modeling semantic shifts.
\textit{Video scene segmentation}, often studied on movies \cite{huang2020movienet}, is primarily focusing on grouping
scenes with similar content~\cite{Rotman2017RobustVS,rao2020local, huang2020movienet,chen2021shot,pardo2022moviecuts,Mun2022BaSSL,Wei2023multimodal, Sadoughi2023Mega,chen2023movies2scenes,islam2023efficient,Yang2023SceneSegmentation,Park2024ContrastingMS}.
Another line of work considers boundary detection for
\textit{temporal action segmentation}~\cite{behrmann2022unified,farha2019ms,gao2021global,li2021temporal,yi2021asformer},
or localization \cite{cheng2022tallformer,liu2022end,zhang2022actionformer,zeng2019graph}.
Unlike chaptering with free-form text, action segmentation assigns a label from a predefined set of categories, and typically
defines short atomic actions as the unit.
In contrast to these tasks, chapter boundaries can take various different forms depending on the type and the granularity of the video
(e.g., each exercise within sports video, each slide within a lecture, each step in instructional video, each topic in a podcast video).
Shot, scene, or action boundaries therefore may or may not correspond to complex chapter boundary definitions.
Moreover, these tasks are mostly tackled with vision-only inputs~\cite{sidiropoulos2011temporal,yi2021asformer,zhang2022actionformer},
\new{without leveraging speech.} %
While text and audio segmentation have also been tackled separately~\cite{retkowski2024textchaptering, Ghazimatin_2024}, video chaptering is based on both audio and vision inputs~\cite{yang2023vidchapters}.

\vspace{-0.2cm}
\paragraph{Video captioning.} %
Generating chapter titles \cite{yang2023vidchapters} is relevant to the task of captioning that seeks to describe the video content with text.
There is a large literature on single video captioning~\cite{lin2022swinbert,seo2022end,shvetsova2023howtocaption,chen2024vast},
often focusing on short video clips. Typical datasets for training
such as %
MSR-VTT~\cite{xu16msrvtt}, WebVid~\cite{bain2021frozen}, HowTo100M~\cite{miech19howto100m}, Video-CC~\cite{nagrani2022learning}
include captions of videos spanning a few seconds (5-15sec on average).
In \textit{generic event boundary captioning}~\cite{wang2022geb}, event intervals are similarly short, in the order of 2 seconds.
On the other hand, \textit{video summarization} methods operate on longer videos;
however, their goal is to reduce the entire video into a single summary description
\cite{zeng2016title,amirian2021automatic,zhang2020comprehensive,Lin2023UniVTG,islam2024videoReCap,argaw2024videosummarization, zhao2024cap2sum,he2023align}, %
not necessarily with a temporal segmentation component. %
\textit{Dense video captioning}~\cite{krishna2017dense,zhou2018end,huang2020multimodal,wang2021end,yang2023vid2seq,Zhou2024Streaming}
is the closest to video chaptering in terms of problem formulation,
aiming to both temporally localize and caption different events.
Indeed, prior work on video chaptering trains the dense captioning method of Vid2Seq~\cite{yang2023vid2seq}
on the VidChapters-7M dataset~\cite{yang2023vidchapters},
\new{but relies on a fixed number of equally sampled frames.} 
In this paper, we leverage some of the annotations of this dataset to train
an LLM-based chaptering model substantially outperforming previous methods~\cite{yang2023vid2seq,yang2023vidchapters}.

\vspace{-0.2cm}
\paragraph{Long-form video understanding.}
The definition of long videos has evolved 
with the release of various datasets spanning
seconds \cite{xue2022advancing,xiao2021next}, %
a few minutes \cite{sun2022long,mangalam2023egoschema,Ego4D2022CVPR,fang2024mmbenchvideo}, %
10-30 minutes \cite{argaw2024videosummarization,MLVU}, %
or one hour \cite{Wu2023NewsNet,yang2023vidchapters,soldan2022mad,fu2024mme,islam2024videoReCap}. %
MLVU~\cite{MLVU} introduces a benchmark for evaluating multiple long video understanding tasks such as summarization and QA; however, the data is not suitable for chaptering due to lack of annotations. Video-MME~\cite{fu2024mme} also contains hour-long videos for QA.
MAD~\cite{soldan2022mad,han23_autoad} provides audio description for long movies,
but each description spans a few seconds and the sparse coverage over the video is different from contiguous chapters.
Recently, Ego4D-HCap~\cite{islam2024videoReCap} was proposed for hierarchical video summarization.
However, this dataset involves dense captioning with visual inputs only, while we focus on video chaptering with visual and speech inputs.
To the best of our knowledge, VidChapters-7M~\cite{yang2023vidchapters} is the only open-sourced dataset for training and evaluating chapter generation,
which we employ in this paper.
Non-public related datasets include 
NewsNet~\cite{Wu2023NewsNet} which includes hierarchical temporal segmentation annotations,
the TV news chaptering dataset used in \cite{guetari2024}, and the ChapterGen dataset \cite{cao2022multi}.

Increased video lengths led to a range of works focusing on efficient temporal modeling strategies.
A common technique to deal with longer videos is to use pre-extracted visual features \cite{zellers2021merlot,soldan2022mad,han23_autoad}.
For end-to-end learning with transformers,
several works
explored factorized spatio-temporal attention \cite{bertasius202timesformer,arnab2021vivit,bain2021frozen}. %
Others have looked at various ways to incorporate
memory mechanisms \cite{Wu_2022_CVPR,Kahatapitiya2024langrepo},
blockwise attention~\cite{liu2023ring,liu2024worldmodel}, or captioning frames to exploit LLMs 
\cite{zhang2024simplellm,wang2024videotree}.
Given the redundancy in consecutive video frames, frame selection methods were
explored in the context of short video captioning and action recognition \cite{Wu_2019_CVPR,Chen_2018_ECCV},
as well as `long' video QA in 3-minute durations \cite{yu2023VideoQA,Tan2024Koala,Papalampidi_2024_CVPR}.
Most common approach with current large video models is to perform sparse sampling
with equal spacing \cite{lei2021less,yang2023vid2seq,chen2023grounding}.
SCSampler~\cite{korbar2019scsampler} exploits the low-dimensional audio modality to
efficiently select salient video clips for action recognition.
In our method, we also leverage audio, but in the form of ASR, and
run the costly frame captioning step only on keyframes on locations
predicted by a speech-based frame selection module.

\vspace{-0.2cm}
\paragraph{LLM use in video understanding.}
LLMs such as GPT~\cite{gpt18openai,GPT_3},
Llama~\cite{touvron2023llama,touvron2023llama2,dubey2024llama}, and Gemini~\cite{geminiteam2024,gemini1p5team2024},
have been leveraged in different ways for improving video understanding.
A popular approach is to train `bridge' modules between pretrained visual backbones \cite{radford2021learning}
and LLMs to build vision-language models (VLMs) that can ingest videos (e.g., Video-Llama~\cite{videollama2023}, Video-LLaVa~\cite{lin2023video}).
Other works have employed LLMs for
automatic construction of video datasets~\cite{islam2024videoReCap,argaw2024videosummarization,ventura24covr2,shvetsova2023howtocaption},
tool use~\cite{min2024morevqa}, storing memory in video QA \cite{Kahatapitiya2024langrepo},
\new{and %
	temporal localization~\cite{huang2024lita}.}
Similar to us, VideoTree~\cite{wang2024videotree} and \new{VideoAgent~\cite{fan2025videoagent}}
caption keyframes before passing them to an LLM together with a question for answer generation,
addressing the limitations of \cite{zhang2024simplellm} which performs a similar methodology without keyframe selection on shorter videos.
In this study, we find that captioning alone is not sufficient, and needs to be complemented with ASR for competitive chaptering performance.
Close to us, \cite{argaw2024videosummarization} exploits ASR on long videos and summarizes them with LLMs to generate pseudo-labels for
video summarization training.
In our work, we leverage LLMs, specifically finetuning a Llama model \cite{dubey2024llama} for chaptering by prompting with speech transcription and frame captions.
We show that finetuning is essential for adapting to the task so that the LLM picks up relevant content within the large context input \cite{shi23}.

    \section{\method:~LLM-based Video Chaptering}
\label{sec:method}

\begin{figure*}
	\centering
	\includegraphics[width=.99\linewidth]{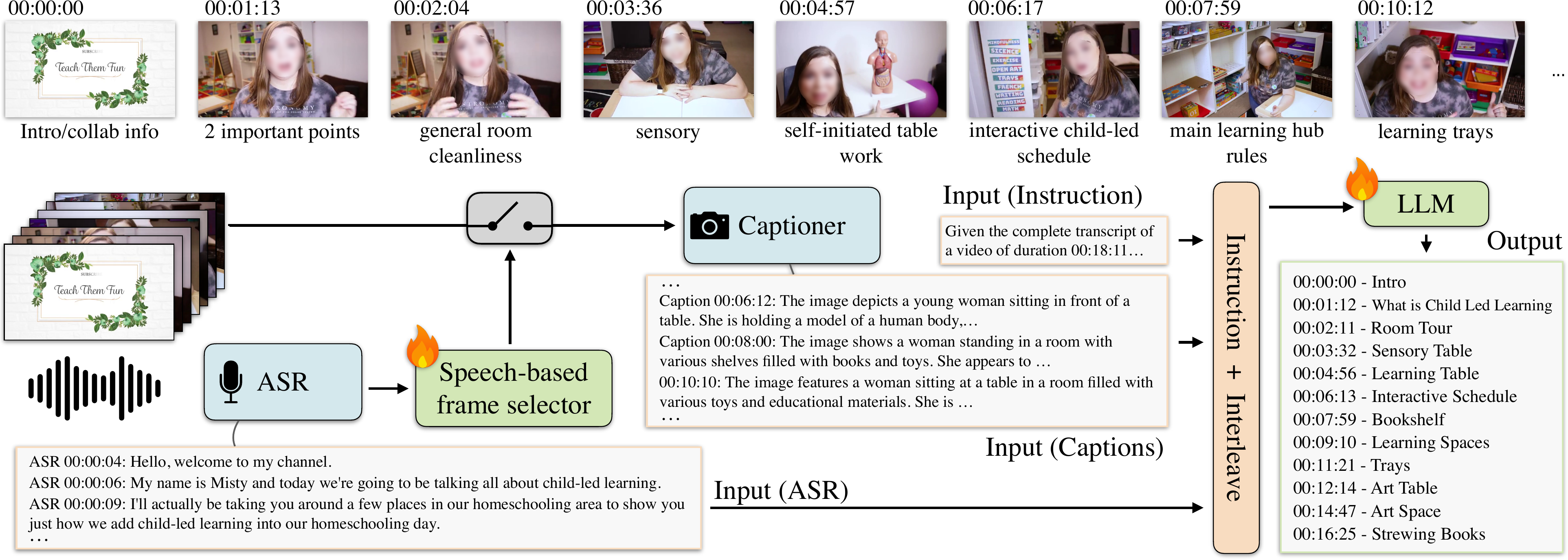}
    \vspace{-2mm}
	\caption{\textbf{Method overview:} Our \method framework first selects video frames to process using speech information. Then we use a visual captioner to map the selected frames in the text space. We feed the resulting captions, along with speech transcripts, to the LLM which outputs the chapter boundaries and titles jointly as a single sequence of tokens.}
	\vspace{-2mm}
	\label{fig:method}
\end{figure*}

We provide an overview of our video chaptering framework, referred to as \method, in \cref{fig:method}.
Given video frames and speech transcripts, we aim at predicting relevant chapter boundaries and titles.
For this, we first select video frames to process with a speech-based frame selection module.
Then we use an off-the-shelf visual captioner to map the selected frames in the text space.
We feed the resulting captions, along with speech transcripts, to the LLM which outputs the chapter boundaries and titles jointly as a single sequence of tokens.
Finally, we devise an iterative prediction procedure in case the input text sequence is too long to handle for the LLM.
We next describe in more detail each component.

\vspace{-2mm}
\paragraph{Task formulation.}
Video chaptering \cite{yang2023vidchapters} aims at segmenting a video into semantically meaningful chapters, and generating a title for each segment.
The chapters are contiguous, with no gaps between them, and together span the entire video duration from start to end.
Formally, given video frames $V = (v_1, v_2, \ldots, v_N)$ 
and temporally-aligned speech transcripts $S = (s_1, s_2, \ldots, s_M)$, where each speech transcript contains an utterance and its associated start and end timestamps,
the task is to output a sequence of chapters $C = (c_1, c_2, \ldots, c_L)$,
where each chapter $c_i$ is a tuple $(b_i, t_i)$ containing a start timestamp $b_i$ and a descriptive title $t_i$. 
The end time of chapter $i$ is implicitly defined by the start time of the subsequent chapter $b_{i+1}$, or total video duration if $i=L$.

\vspace{-2mm}
\paragraph{Speech-based frame selection.}
Video chaptering involves processing hour-long videos. 
Therefore, densely sampling frames is computationally intractable due to numerous inference passes through a vision model (e.g., a visual captioner) and exceeding standard LLM context lengths.
Upon inspection of our data, we found that while the speech transcription has 257 tokens per minute on average,
a caption is 66 tokens long on average hence captions would take 3,960 tokens per minute when sampling a video at 1 FPS.
To address these challenges, 
we employ a frame selection strategy.

Specifically, we use speech transcripts to guide which video frames to process for the vision model.
This is done by first training a speech-only variant of our LLM
to predict a sequence of chapter boundaries $\{\hat{b}_1, \hat{b}_2, ..., \hat{b}_K\}$ from speech transcripts $S$ only. For each predicted boundary $\hat{b}_i$, we sample a frame $v_i$ from the video at that timestamp.
Note that this variant is cheaper compared to the full model 
as it only
\new{needs ASR transcription from the audio stream, %
without requiring any processing of the RGB %
stream (i.e., captioning).} %
We then process the video frames \new{only} at the time locations predicted by this model.
The visual information thus complements the previous `blind' predictions from the narrations,
and allows us to refine the predictions.
This results in a video representation $V_{sampled} = (v_1, v_2, \ldots, v_K)$ where $K<<N$.
For the videos that lack speech entirely (e.g., about 3\% of the videos in \cite{yang2023vidchapters}), 
we sample frames at 10-second intervals, with an upper bound of 100 frames to maintain computational practicality.

\vspace{-2mm}
\paragraph{Mapping video to text with timestamps.}
To leverage the knowledge of a pretrained LLM, we map all our inputs to text.
This includes:
(1)~speech transcriptions $S = (s_1, s_2, \ldots, s_M)$ from the audio modality, and
(2)~caption descriptions $V_{captions} = (d_1, d_2, \ldots, d_K)$ from the visual modality.
In detail, for speech transcriptions, we use ASR outputs provided by~\cite{yang2023vidchapters},
obtained using the Whisper-Large-V2~\cite{radford2023whisper} model
through the WhisperX~\cite{bain2022whisperx} implementation.
For captioning, we employ MiniCPM-V~\cite{yao2024minicpm}
as an image captioner, applied independently on the selected video frames, i.e., $d_i = Captioner(v_i)$.

As we aim at predicting relevant chapter boundaries, we provide temporal information to the LLM.
For both modalities, we prepend the timestamp information
formatted as ``\texttt{HH:MM:SS}''
to encode the location at which the speech or caption is obtained.

Captions naturally come from a single point in time.
Speech segments cover intervals, but their duration is typically very short ({3-4 seconds}).
We therefore simply use the start time of each transcribed speech interval.
We interleave the speech and caption inputs based on their timestamps in a sorted order.
We add a modality-specific prefix to each timestamp to denote which modality the information is extracted from
(i.e., \texttt{ASR} for speech transcripts, \texttt{Caption} for captions).

We prepend the text combining speech transcripts and captions with a fixed prompt that provides task instructions (see sup.~mat.~for the exact wording).
This prompt occupies approximately 90 tokens and is independent of video length.

\vspace{-2mm}
\paragraph{Language model.}
We derive our framework by making use of a powerful pretrained LLM.
Specifically, we employ the recent
Llama-3.1-8B-Instruct~\cite{dubey2024llama} model
and further finetune on chapter annotations using the LoRA technique~\cite{hu2022lora}.
Given the input structure previously described, the LLM
is trained to output chapters, 
where each chapter consists of a timestamp in \texttt{HH:MM:SS} format followed by a free-form 
chapter title. 
We treat both the timestamps and titles simply as text tokens and apply the standard cross-entropy loss over the original vocabulary of the pretrained LLM. %
We apply teacher forcing during training and decode tokens autoregressively at inference.
\new{Note that the final model (taking both speech and captions as input) is trained independently
from the speech-only version of our model used for frame selection,
but these two models share the same backbone,
and only differ in their LoRA parameters (13MB each).}
Across all experiments, we finetune models for a single epoch and use the same hyperparameters.
We provide these hyperparameters, 
along with implementation details in
\appendixref{Appendix~A}{\cref{sec:app:implementation}},
\new{and provide experiments with several Llama variants}
in
\appendixref{Appendix~C}{\cref{sec:app:additional-experiments}}.

\vspace{-2mm}
\paragraph{Iterative prediction for long videos.}
The inputs may exceed the context window limitation of the LLM, especially in the case of long videos.
For example, 
on an A6000 GPU, the Llama-3.1-8B-Instruct~\cite{dubey2024llama} model can process videos up to around 15k tokens during training, which corresponds to 50 minutes of video content on average, and 25k tokens during inference, which corresponds to 80 minutes of video content on average.
To address this issue, during training, 
we select videos that have less than 15k tokens.
Since there are videos up to 1 hour long in the training set that satisfy this constraint, and since we do not need the entire training dataset to achieve good performance, this token limitation does not hinder our training.
During evaluation, we predict chapters for each chunk sequentially, such that the start of a chunk is the end of the previous chunk.
Finally, we merge the predictions from all chunks to obtain chapter boundaries for the complete video.
\new{We provide more details in 
\appendixref{Appendix~A.4}{\cref{sec:app:implementation:iterative_prediction}}.}

    \section{Experiments}
\label{sec:experiments}

In this section,
we start by describing the data and evaluation metrics used in our experiments (\cref{subsec:data-evaluation}).
Next, we compare our results with the state of the art (\cref{subsec:sota}),
and then provide a series of
ablations in our framework (\cref{subsec:ablation}).
Finally, we investigate the impact of testing with very long videos
exceeding our context window limitations (\cref{subsec:longer-videos}).

\subsection{Data and evaluation}
\label{subsec:data-evaluation}

\paragraph{Data.}
We train and evaluate on the recently released VidChapters-7M~\cite{yang2023vidchapters} dataset that includes user-annotated chaptered videos sourced from YouTube.
Speech transcripts are obtained using Whisper~\cite{radford2023whisper} as the ASR method.
In the original release, there is a total of 817k videos, spanning 8M chapters, with
2.4 minutes per chapter and 5.4 words per chapter title, totaling to 23 minutes and 8.3 chapters per video on average.
Data is split into 801k training, 8.2k validation, and 8.2k test videos.
To measure performance at different video lengths, we define three categories depending on video duration:
`short' (0-15min), `medium' (15-30min), and `long' (30-60min) videos. 
In this work, we use a subset of the training data as we observe increasing the training set brings diminishing returns at the cost of extended training times (see \cref{fig:datasize}).
Specifically, we use about \new{20k} training videos
\new{(10k short videos used for the speech-based frame selection model
and another 10k videos evenly split across short, medium and long durations for the final model).}
For state-of-the-art comparisons (\cref{subsec:sota}), we 
employ the full official test set, which also contains videos without any speech (2.5\% of the videos),
and videos longer than 60 minutes (e.g., there are few videos that last about 12 hours).
In ablations (\cref{subsec:ablation}),
both for faster experimentation, and
to limit the use of the test set during experimentation,
we train on a randomly sampled subset of 1k videos (\new{evenly split between short, medium, and long}) and
report results on a randomly sampled subset of \new{300} \textit{validation} videos
(\new{100 from each duration}) that have at least one speech utterance.

\vspace{-2mm}
\paragraph{Evaluation metrics.}
We primarily monitor temporal segmentation metrics to evaluate our chapter boundary detections.
In particular, we employ \textbf{tIoU} and \textbf{F1} scores.
For tIoU (temporal Intersection over Union),
we first compute the optimal matching between predicted and ground truth segments by greedily selecting pairs with highest IoU scores.
The tIoU score is then calculated as the mean IoU across all matched pairs, multiplied by 100 to obtain a percentage.
For F1 score, 
we first compute precision and recall at different IoU thresholds (ranging from 0.5 to 0.95 with a step of 0.05).
At each threshold,
a prediction is considered correct if it has IoU above the threshold with a ground truth segment.
The precision is the ratio of correct predictions to total predictions,
while recall is the ratio of matched ground truth segments to total ground truth segments.
The F1 score is then computed as the harmonic mean of precision and recall.
The final F1 metric is the average across all thresholds,
multiplied by 100 to obtain a percentage.
Note that \cite{yang2023vidchapters} uses recall and precision metrics
in two ways: (1) by considering timestamps within 3 or 5 second thresholds as matches,
and (2) by considering segments with IoU above 0.5 or 0.7 as matches.
While these metrics provide point estimates at specific thresholds,
we find that tIoU and F1 scores offer several advantages:
they evaluate performance continuously across multiple thresholds,
are more interpretable,
and provide a more comprehensive evaluation of the model.
For completeness, we also report the metrics used in \cite{yang2023vidchapters}
\new{in \appendixref{Appendix~C}{\cref{sec:app:additional-experiments}}.}

For chapter title evaluation, we follow \cite{yang2023vidchapters} and report
\textbf{SODA (S)}~\cite{fujita2020soda} and \textbf{CIDEr (C)}~\cite{vedantam2015cider},
which measure the quality of the titles for the predicted segments
that match to the ground segments
(see \cite{yang2023vidchapters}
for details).

\begin{table*}
  \centering
  \setlength\tabcolsep{4pt}
  \resizebox{0.99\linewidth}{!}
  {
  \begin{tabular}{@{}llc|cccc|cccc|cccc|cccc@{}}
    \toprule
    & Frame & & \multicolumn{4}{c|}{Short} & \multicolumn{4}{c|}{Medium} & \multicolumn{4}{c|}{Long} & \multicolumn{4}{c}{All}\\
    Backbone & selection & Ft. & F1 & tIoU & S & C & F1 & tIoU & S & C & F1 & tIoU & S & C & F1 & tIoU & S & C \\
    \midrule
    \gray{GPT-4o-mini~\cite{openai2024gpt4ocard}}$\dagger$ & \gray{Ours} & \gray{\xmark}
      & \gray{32.1} & \gray{64.5} & \z \gray{7.2} & \z \gray{42.4}
      & \gray{30.5} & \gray{62.3} & \z \gray{6.1} & \gray{30.6}
      & \gray{28.0} & \gray{61.0} & \z \gray{6.0} & \gray{27.3}
      & \gray{31.2} & \gray{63.6} & \z \gray{6.8} & \z \gray{37.8}\\
    \gray{GPT-4o~\cite{openai2024gpt4ocard}}$\dagger$ & \gray{Ours} & \gray{\xmark}
      & \gray{37.7} & \gray{68.0} & \z \gray{8.4} & \z \gray{53.8}
      & \gray{38.1} & \gray{68.8} & \z \gray{8.1} & \gray{51.4}
      & \gray{36.5} & \gray{66.2} & \z \gray{6.6} & \gray{34.8}
      & \gray{37.6} & \gray{68.0} & \z \gray{8.1} & \z \gray{51.0} \\
    \gray{Gemini-2.0-Flash~\cite{geminiteam2024}}$\dagger$ & \gray{Ours} & \gray{\xmark} 
      & \gray{39.9} & \gray{69.2} & \gray{12.0} & \z \gray{72.8}
      & \gray{43.8} & \gray{71.4} & \gray{11.2} & \gray{70.3}
      & \gray{34.9} & \gray{66.2} & \gray{\z 9.0} & \gray{51.6}
      & \gray{40.2} & \gray{69.3} & \gray{11.4} & \gray{\z 69.7}  \\
    \gray{Gemini-1.5-Pro~\cite{geminiteam2024}}$\dagger$ & \gray{Ours} & \gray{\xmark} 
      & \gray{41.7} & \gray{70.6} & \gray{11.7} & \z \gray{65.3}
      & \gray{43.8} & \gray{71.8} & \gray{11.2} & \gray{61.4}
      & \gray{\textbf{41.3}} & \gray{\textbf{70.6}} & \gray{10.1} & \gray{55.3}
      & \gray{42.2} & \gray{70.9} & \gray{11.4} & \gray{\z 63.2}\\
    \midrule
    Vid2Seq~\cite{yang2023vid2seq,yang2023vidchapters} & Equidistant & \xmark 
    & \z 2.5 & 28.6 & \z 0.3 & \z \z 0.3 
    & \z 3.2 & 29.7 & \z 0.3 & \z 0.4 
    & \z 4.6 & 32.0 & \z 0.3 & \z 0.5 
    & \z 3.0 & 29.3 & \z 0.3 & \z \z  0.4\\
  Llama 3.1-8B & Ours & \xmark
    & 29.9 & 63.4 & \z 7.1 & \z 34.5
    & 30.6 & 62.7 & \z 5.4 & 28.1
    & 26.6 & 59.3 & \z 3.6 & 18.9 
    & 29.5 & 62.5 & \z 6.2 & \z 30.7 \\
    \midrule
    Vid2Seq~\cite{yang2023vid2seq,yang2023vidchapters} & Equidistant & \cmark 
      & 33.4 & 63.7 & 15.2 & \z 74.9 
      & 19.0 & 53.3 & \z 7.5 & 31.9 
      & 16.7 & 50.8 & \z 5.9 & 28.4 
      & 26.7 & 58.6 & 11.6 & \z 55.8 \\
    Llama 3.1-8B (\method) & Ours & \cmark 
      & \textbf{45.5} & \textbf{72.2} & \textbf{20.2} & \textbf{103.5}
      & \textbf{46.7} & \textbf{72.3} & \textbf{18.8} & \textbf{98.7}
      & \textbf{41.3} & {69.2} & \textbf{15.8} & \textbf{91.2}
      & \textbf{45.3} & \textbf{71.8} & \textbf{19.3} & \textbf{100.9} \\
    \bottomrule
  \end{tabular}
  }
\vspace{-0.2cm}
  \caption{\textbf{Comparison to the state of the art on VidChapters-7M test set:}
  	We split the table into (bottom) the comparison between \method and the
  	state-of-the-art
  	method Vid2Seq~\cite{yang2023vid2seq},
  	and
  	(top) the evaluation of \gray{proprietary models}.
  	\method significantly outperforms Vid2Seq trained and reported by~\cite{yang2023vidchapters} (45.3 vs 26.7 F1).
  	Our method also achieves strong performance in zero-shot mode -- without finetuning \new{(Ft.)} on any chapter annotation (29.5 F1).
  	\new{Furthermore, we report performance
  	of proprietary models in such zero-shot setting,
  	using our speech-based frame selection and captioning, and observe inferior results than \method (42.2 F1 with Gemini-1.5-Pro).}
  Note that we use the full official 8.1k test set videos (`All'), unlike in the remaining experiments
  that report on the validation subset. We also report the performance breakdown into short (4891),
  medium (1736), and long (892) test videos. 
 Our model was trained on 10k  \new{videos balanced across short, medium and long durations.}
  \new{$\dagger$ denotes evaluation on a random 10\% subset of the test set due to API costs of proprietary models.}
  }
\vspace{-4mm}
  \label{tab:sota}
\end{table*}

\begin{figure*}
	\centering
	\includegraphics[width=.99\linewidth]{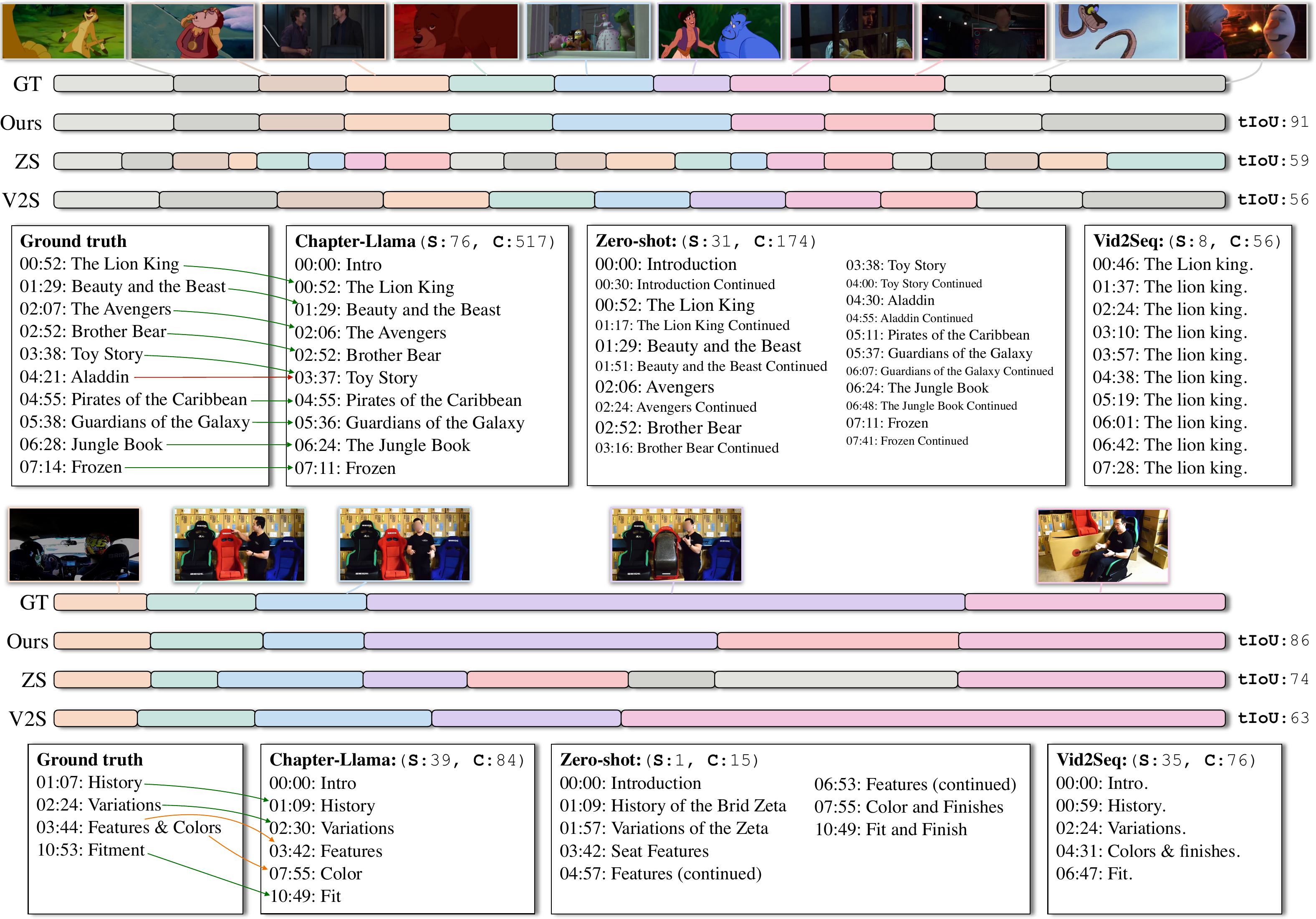} 
	\vspace{-0.2cm}
	\caption{
		\textbf{Qualitative results:} We display two examples
		and compare our \method results against the ground truth (GT),
		as well as the zero-shot (ZS) and Vid2Seq (VS) baselines.
		\new{For each example, we show the corresponding SODA (S) and CIDEr (C) scores.}
		Our method overall shows the highest similarity with the GT,
		while Vid2Seq can suffer from repeated chapter titles, and
		zero-shot generations tend to over-segment.
	}
	\label{fig:qualitative}
	\vspace{-2mm}
\end{figure*}

\subsection{Comparison with the state of the art}
\label{subsec:sota}

In \cref{tab:sota}, we report the performance of our model on the full VidChapters-7M test set
\cite{yang2023vidchapters} (`All' columns),
and compare to the state of the art reported in \cite{yang2023vidchapters},
which uses Vid2Seq~\cite{yang2023vid2seq}.
\new{Moreover, we evaluate four proprietary models 
using our %
speech-based frame selection and captioning in a zero-shot manner.

We observe that our finetuned \method achieves substantial performance improvements
across all metrics and video duration categories.
(e.g., 45.3 vs 26.7 F1 and 19.3 vs 11.6 SODA compared to Vid2Seq).
}
Notably, our improvement 
\new{over Vid2Seq}
is more important for medium and long videos compared to short videos.
Note that our final approach was trained using the subset of data detailed in the previous section, 
specifically 20k videos, which constitutes only 2.5\% of the total available 
training data.
In contrast, the baseline Vid2Seq model~\cite{yang2023vid2seq} 
was trained on a considerably larger dataset, 
utilizing both HowTo100M~\cite{miech19howto100m} 
and the entire VidChapters-7M training set.

Additionally, we report performances of our model without training
on any chapter annotations 
(i.e., both the speech-based frame selector and the LLM are not finetuned, and run with the same prompt as in the finetuned setting).
We see that our zero-shot method
also achieves competitive performance (e.g., 29.5 F1), whereas Vid2Seq
only trained on HowTo100M does not generalize (3.0 F1).

\new{Finally, when zero-shot evaluating the proprietary models,
GPT4-o~\cite{openai2024gpt4ocard} and 
Gemini variants~\cite{geminiteam2024},
with our speech-based frame selection and captioning inputs,
we observe competitive performances (e.g., 42.2 F1 with Gemini-1.5-Pro);
however, our \method still surpasses on all metrics.
Note that, due to API costs of the proprietary models,
we performed their evaluation on a random 10\% subset 
of the test set; however, we verified that the scores
are similar between 10\% and 100\% of the test set when
evaluating with \method.}

\vspace{-2mm}
\paragraph{Qualitative comparison.}
In \cref{fig:qualitative}, 
we provide qualitative examples comparing our method against Vid2Seq~\cite{yang2023vid2seq,yang2023vidchapters} and our zero-shot baseline. 
Our predictions align well with the ground truth chapters, 
accurately capturing both the temporal boundaries 
and generating relevant titles. %
In contrast, Vid2Seq segments tend to be less accurate, and
we also observe that it
often produces repetitive titles (bottom example).
The zero-shot \method baseline tends to generate relatively longer and verbose chapter titles
and often generates chapters that appear to be continuations of previous chapters rather than distinct segments, while also exhibiting over-segmentation issues.
We provide more examples in 
\new{\appendixref{Appendix~E}{\cref{sec:app:qualitative-analysis}}.}

\subsection{Ablation studies}
\label{subsec:ablation}
In the following, we experiment with %
(i)~the contribution of speech and caption modalities, along with the effect of LLM finetuning, %
(ii)~the effect of our frame selection method for captioning, %
(iii)~the amount of training data, %
and
\new{(iv)~the use of frame embeddings instead of captions}. %
As mentioned above, we use 1k training and \new{300} validation videos
for these ablations.

\begin{table}
    \centering
    \resizebox{0.9\linewidth}{!}
    {
    \begin{tabular}{ccc|cc|ccc}   
    \toprule
    & \multicolumn{2}{c|}{Modalities} & \multicolumn{2}{c|}{Segmentation} & \multicolumn{2}{c}{Titles}  \\
    \midrule
    & Speech & Captions & F1 & tIoU & S & C \\
    \midrule
    \multirow{3}{*}{\begin{tabular}[c]{@{}c@{}}\rotatebox{90}{{\small Zero-shot}}\end{tabular}}
    & \xmark & \cmark & 12.6 & 48.6 & \z 1.9 & \z 6.4 \\
    & \cmark & \xmark & 22.7 & 57.3 & \z 4.4 & 19.7 \\
    & \cmark & \cmark & 29.9 & 63.0 & \z 6.9 & 33.7 \\
    \midrule
    \multirow{3}{*}{\begin{tabular}[c]{@{}c@{}}\rotatebox{90}{{\small Finetuned}}\end{tabular}}
    & \xmark & \cmark & 39.1 & 67.7 & \z 5.9 & \z 20.2 \\
    & \cmark & \xmark & 38.5 & 68.1 & 13.9 & \z 67.3 \\
    & \cmark & \cmark & \textbf{42.6} & \textbf{70.6} & \textbf{16.4} & \z \textbf{82.4} \\
    \bottomrule
    \end{tabular}
    }
\vspace{-0.2cm}
    \caption{\textbf{Contribution of different modalities and finetuning:} 
    Finetuning the LLM with 1k videos largely improves chaptering performance on \new{300} validation videos, see bottom block vs top block.
    In the finetuned setting, 
    we further demonstrate the advantages of
    combining both modalities, i.e., transcribed speech from ASR and automatic captions
    extracted from video frames.
    }
\vspace{-0.3cm}
    \label{tab:modalities}
\end{table}

\vspace{-2mm}
\paragraph{Modalities and LLM finetuning.}
\label{subsec:modalities}
In \cref{tab:modalities},
we ablate the impact of finetuning the LLM and the contribution of each of the speech and caption modalities. 
In the top block, we run our baselines in zero-shot setting as introduced in the previous section.
The speech-only baseline outperforms the captions-only baseline by a large margin in the zero-shot setting. 
This suggests that speech contains more relevant information for chaptering, as was previously
observed by \cite{yang2023vidchapters}.

As shown in the bottom block of \cref{tab:modalities},
we observe large performance improvements when finetuning the LLM,
as opposed to zero-shot.
We hypothesize that zero-shot prompting with a long multi-modal text, potentially containing
redundant and irrelevant information, may overwhelm the LLM \cite{shi23,wang2024videotree}.
We obtain our best model by combining the two modalities, which performs better than the individual speech-only or caption-only models.
This demonstrates the multi-modal capabilities of our model.

\begin{table}
    \centering
    \setlength\tabcolsep{3pt}
    \resizebox{1\linewidth}{!}
    {
    \begin{tabular}{cclcc|cc|cc}   
    \toprule
    \multicolumn{2}{c}{\multirow{2}{*}{Method}} & Frame selection & \new{average} & \#tokens & \multicolumn{2}{c|}{Segmentation} & \multicolumn{2}{c}{Titles} \\
    & & for captions & \new{\#frames} & per min. & F1 & tIoU & S & C \\
    \midrule
    \rowcolor{lightgray}\multicolumn{2}{l}{\textsc{Baselines}} & & & & & & & \\
    \multicolumn{2}{l}{Shot detection~\cite{brandon2018}} 
        & n/a & \z 49.4 & n/a & 6.2 & 37.6 & - & - \\
    \multicolumn{2}{l}{Vid2Seq~\cite{yang2023vid2seq,yang2023vidchapters}} 
        & 100 equidistant & 100.0 & 128.6 & 25.4 & 57.8 & 11.2 & 55.0  \\
    \midrule
    \rowcolor{lightgray}\multicolumn{3}{l}{\textsc{\method Variants}} & & & & & & \\
    Speech & Caption & & & & & & \\
        \cmark & \xmark & n/a & n/a & 248.6 & 38.5 & 68.1 & 13.9 & 67.3  \\
    \cmidrule(lr){1-9} 
    \multirow{5.25}{*}{\begin{tabular}[c]{@{}c@{}}{\xmark}\end{tabular}} 
    & \multirow{5.25}{*}{\begin{tabular}[c]{@{}c@{}}{\cmark}\end{tabular}} 
        & 100 equidistant & 100.0 & 449.1 & 21.0 & 53.8 & \z 8.4 & 36.0 \\
        & & Every 10 sec. & \z 83.1 & 280.3 & 12.8 & 45.9 & \z 4.3 & 13.0\\
        & & \new{Shot boundaries} & \z 49.4 & 193.2 & 16.2 & 50.7 & \z 3.9 & 12.4 \\
        & & \new{10 equidistant} & \z 10.0 & \z 41.8 & 11.0 & 46.4 & \z 3.6 & \z 9.0 \\
        & & Speech-based & \z 10.3 & \z 36.2 & 39.1 & 67.7 & \z 5.9 & 20.2 \\
    \cmidrule(lr){1-9} 
    \multirow{5.25}{*}{\begin{tabular}[c]{@{}c@{}}{\cmark}\end{tabular}} 
    & \multirow{5.25}{*}{\begin{tabular}[c]{@{}c@{}}{\cmark}\end{tabular}} 
        & 100 equidistant & 100.0 & 746.2 & 39.2 & 67.4 & 16.1 & \textbf{83.8} \\
        & & Every 10 sec. & \z 83.1 & 570.1 & 41.0 & 69.3 & 15.4 & 77.3 \\
        & & \new{Shot boundaries} & \z 40.4 & 481.7 & 40.6 & 69.1 & 15.8 & 79.3 \\
        & & \new{10 equidistant} & \z 10.0 & 326.1 & 40.1 & 67.9 & 15.8 & 77.5\\
        & & Speech-based &  \z 10.3 & 320.4 & \textbf{42.6} & \textbf{70.6} & \textbf{16.4} & {82.4} \\
    \bottomrule
    \end{tabular}
    }
\vspace{-0.2cm}
    \caption{\textbf{Frame selection strategies for captioning:} 
    We evaluate different approaches for selecting frames to extract captions from, 
    comparing our speech-based selection method against baselines.
    The table shows results for models trained on 1k videos and evaluated on \new{300} validation videos.
    We experiment with using speech only, captions only, and both modalities (bottom section).
    For caption extraction,
    we compare \new{our speech-based approach to other alternatives
    such as equidistant sampling (100 or 10 frames),
    uniformly sampling every 10 seconds,
    or sampling at shot boundaries using~\cite{brandon2018}.}
    Our speech-based frame selection achieves the best overall performance (\new{F1: 42.6, tIoU: 70.6}) while requiring significantly fewer \new{number of frames on average (10.3)} compared to
    \new{other}
    sampling approaches.
    The tokens-per-minute statistic %
    shows the total input length including both speech transcriptions and captions, excluding the fixed prompt template.
    }
    \vspace{-0.3cm}
    \label{tab:frame_selection}
\end{table}

\vspace{-2mm}
\paragraph{Speech-based frame selection.}
\label{subsec:frame_selection}
In \cref{tab:frame_selection},
we examine a number of strategies to sample frames
at which we extract captions.
In addition to previously described metrics,
for each of the frame sampling approaches,
we report 
\new{the average number of captions per video}
and the average number of \new{text} tokens per minute.
For reference, we also report an off-the-shelf shot detection~\cite{brandon2018} and Vid2Seq~\cite{yang2023vid2seq,yang2023vidchapters}. 

We compare our speech-based frame selection strategy to various baselines.
We experiment with sampling 
(i)~uniformly 100 frames as in Vid2Seq, %
(ii)~every 10 seconds,
\new{(iii)~at shot boundaries detected by an off-the-shelf shot detector~\cite{brandon2018},
(iv)~10~equidistant frames to be similar to our speech-based locations (i.e., 10.0 vs 10.3 number of frames on average),}
and (v)~sampling at frames predicted as chapter boundaries by our LLM that inputs only speech.
\new{In all cases, we limit the maximum number of frames to 100 as in \cite{yang2023vid2seq,yang2023vidchapters} to handle extreme durations.} %

In both caption-only and caption+speech settings, our speech-based frame selection approach achieves better segmentation results than the
\new{more frame-expensive baselines `100 equidistant', `every 10 sec', and `shot boundaries',}
while using much less frames, and also improves over the
\new{`10 equidistant'}
baseline which uses a similar number of frames.
This demonstrates the effectiveness of our speech-based frame selection strategy.

For reference, we also report positive comparison against shot detection and Vid2Seq \cite{yang2023vid2seq,yang2023vidchapters}.
Note Vid2Seq has less \#tokens per min.\ compared to our 100 equidistant variants, because Vid2Seq uses a different timestamp tokenizer in the input.

\begin{figure}
	\centering
	\includegraphics[width=.99\linewidth]{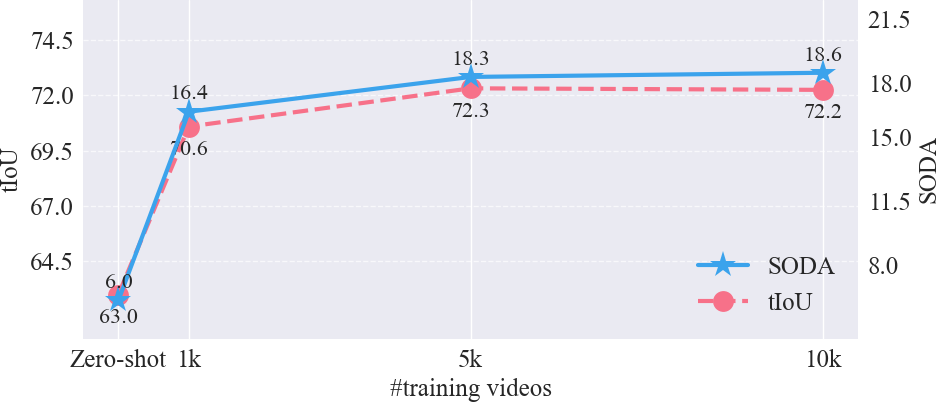}
	\vspace{-0.2cm}
	\caption{\textbf{\new{Amount of training data:}}
		Our experiments show a substantial improvement when moving from zero-shot to training with 1k videos.
		\new{Beyond 1k videos, performance continues to improve but at a much slower rate,}
		motivating our choice of using only 10k training videos for our final LLM.
	}
	\vspace{-3mm}
\label{fig:datasize}
\end{figure}

\vspace{-2mm}
\paragraph{Amount of training data.}
Given the large-scale nature of the VidChapters-7M training set,
we investigate how much chapter data is needed for LoRA finetuning the
LLM. 
We plot the performance against the number of training videos in \cref{fig:datasize}.
We start by the zero-shot baseline as the first data point, and report our method
with 1k, 5k, 7k, and 10k videos, %
\new{split evenly between three durations.} %
We see that after increasing above several thousand training videos starts to bring diminishing returns. 
We therefore keep 10k training videos for our final LLM, which makes our approach highly efficient to train (40min on 4 H100 GPUs).
Note that here we focus on the chaptering LLM and always use frame sampling locations from a speech-based module trained on 10k separate videos.

\vspace{-2mm}
\paragraph{\new{Frame embeddings vs captions.}}
\new{In \cref{tab:modalities_mm},
we investigate whether raw visual embeddings could serve as an alternative to textual captions. 
To this end, we 
experiment with replacing the captions with frame embeddings. 
Specifically, for each frame,
we extract the 1152-dimensional output embedding corresponding to the [CLS] token from a frozen SigLIP model~\cite{zhai2023siglip},
and feed through a 2-layer MLP mapping network. 
We initialize the MLP weights from MANTIS~\cite{jiang2024mantis} 
and train jointly with the LLM during finetuning.
The results with `Speech+Embeddings'
are better than `Speech' alone (38.5 vs 40.4 F1), but worse than `Speech+Captions' (42.6 vs 40.4 F1).
The performance gap between `Speech+Embeddings' and `Speech+Captions' 
may be due to the richer information provided by captions, 
which use \textit{multiple} tokens per frame, directly in \textit{text form},
compared to the \textit{single} [CLS] token frame embedding, requiring a \textit{mapping network} to be ingested by an LLM.
Finally, while combining all modalities achieves the best performance (44.4 F1),
we exclude frame embeddings from our final model due to practical considerations, e.g.,
they add complexity, increase processing time by 2.5x,
and require 3000x more storage space.}

\begin{table}
    \centering
    \setlength\tabcolsep{4pt}
    \resizebox{0.99\linewidth}{!}
    {
    \begin{tabular}{ccc|cc|ccc}   
    \toprule
    \multicolumn{3}{c|}{Modalities} & \multicolumn{2}{c|}{Segmentation} & \multicolumn{2}{c}{Titles}  \\
    \midrule
    Speech & Embeddings & Captions & F1 & tIoU & S & C \\
    \midrule
    \cmark & -  & - 
        & 38.5 & 68.1 & 13.9 & 67.3 \\
    - & \cmark & - 
        & 38.4 & 66.5 & \z 3.4 & \z 7.3 \\
    - & - & \cmark          
        & 39.1 & 67.7 & \z 5.9 & 20.2 \\
    \midrule
    \cmark & \cmark & -
        & 40.4 & 68.2 & 15.3 & 74.9 \\
    \cmark & - & \cmark
        & 42.6 & 70.6 & \textbf{16.4} & 82.4 \\
    \cmark & \cmark & \cmark 
        & \textbf{44.4} & \textbf{71.5} & {16.3} & \textbf{84.2} \\
    \bottomrule
    \end{tabular}
    }
\vspace{-0.2cm}
    \caption{\textbf{\new{Frame embeddings vs captions:}}
    \new{We compare using frame captions versus visual features from a frozen SigLIP model projected through a learned 2-layer MLP mapping network (`Embeddings').
    While
    the `Speech+Embeddings' combination
    performs better than speech alone (40.4 vs 38.5 F1), 
    it underperforms compared to
    the `Speech+Captions' combination (42.6 vs 40.4 F1).
    All models are trained with 1k videos and evaluated on 300 videos.}
    }
\vspace{-0.2cm}
    \label{tab:modalities_mm}
\end{table}

\begin{table} %
\centering
\resizebox{0.9\linewidth}{!}
{
\begin{tabular}{@{}lc|c|cccc@{}}
\toprule
    \multicolumn{2}{c|}{} &avg&  \multicolumn{4}{c}{Subset exceeding 35k tokens} \\
    Window & \# tok. & \# iter. & F1 & tIoU & S & C \\
    \midrule
    \multirow{3}{*}{\begin{tabular}[c]{@{}c@{}}{First}\end{tabular}} 
    & 10k   & 1 & 13.1 & 50.5 & \z 4.0 & 31.2 \\
    & 15k   & 1 & 16.6 & 54.9 & \z 5.4 & 43.3 \\
    & 20k   & 1 & 18.7 & 56.7 & \z 6.6 & \textbf{47.5} \\
    \midrule
    \multirow{3}{*}{\begin{tabular}[c]{@{}c@{}}{Iterative}\end{tabular}} 
    & 10k   & 8.5 &  18.5 & 57.1 & \z 6.9 & 25.1 \\
    & 15k   & 5.4 & 23.6 & 60.1 & \z 8.7 & 35.2 \\
    & 20k   & 4.1 & \textbf{25.3} & \textbf{61.4} & \textbf{10.3} & 44.0 \\
\bottomrule
\end{tabular}
}
\vspace{-0.2cm}
\caption{\textbf{Iterative prediction:} Our iterative prediction procedure improves chaptering results on the subset of 110 videos which exceed 35k tokens compared to the baseline that only runs the LLM once (by only taking the first window, and discarding the rest of the input sequence), across various context windows.
As we increase the context window in the iterative prediction, 
the performance gradually improves and the average number of iterations decreases.
\new{The model is trained with 1k videos.}
}
\vspace{-0.3cm}
\label{tab:window}
\end{table}

\subsection{\new{Iterative prediction} on longer videos}
\label{subsec:longer-videos}
In our ablation studies, \new{our experimental setting considered
training and evaluating with %
videos that fit within the LLM context window.} %
In \cref{tab:window},
we evaluate the benefit of our iterative prediction procedure for handling videos that exceed the LLM context window.
For this, we identify videos in the validation set whose inputs exceed
the LLM inference context window ($>35k$ tokens), resulting in 110 videos.
On this challenging subset, 
we find that our iterative prediction procedure improves 
chaptering results compared to the baseline 
that only runs the LLM once by cropping the input to the first input window,
across various context windows (10k, 15k, and 20k).
We refer to
\new{\appendixref{Appendix~B}{\cref{sec:app:dataset-analysis}}} for details
on the video lengths and statistics of videos that exceed 
the LLM context window.

    \section{Conclusions}
\label{sec:conclusions}
We presented \method, an approach that leverages LLMs for hour-long video chaptering by mapping video to text using speech transcripts and efficiently captioning video frames sampled with a speech-based frame selector.
Our results on VidChapters-7M consequently improved the
state of the art by a large margin.
We experimentally demonstrated the benefits of our components through an extensive ablation study.
One limitation of our approach is that it relies on the accuracy of the ASR and the visual captioner. Future work can explore hierarchical chaptering with several granularities and consider the audio modality beyond speech. 
We also note that the LLM, the visual captioner, and speech transcription models are trained on large Web datasets, which can contain biases
that can lead to inaccurate chaptering, especially for videos depicting
underrepresented topics.

    \paragraph{Acknowledgements.}
This work was granted access to the HPC resources of IDRIS under the 
allocation {2024-AD011014696} made by GENCI.
This work was funded in part by the ANR project 
CorVis ANR-21-CE23-0003-01,
a research gift from Google,
the French government under management of Agence Nationale de la Recherche as part of the “France 2030” program, 
reference ANR-23-IACL-0008 (PR[AI]RIE-PSAI projet),
and the ANR project VideoPredict ANR-21-FAI1-0002-01.
Cordelia Schmid would like to acknowledge the
support by the K\"orber European Science Prize.
The authors would also like to thank 
Guillaume Astruc, %
Nikos Athanasiou, %
Hyolim Kang, %
and Nicolas Dufour
for their feedback.

    {
        \small
        \bibliographystyle{ieeenat_fullname}
        \bibliography{sec/60_references}
    }

    \clearpage
    {\noindent \large \bf {APPENDIX}}\\
    \appendix

\renewcommand{\thefigure}{A.\arabic{figure}} %
\setcounter{figure}{0} 
\renewcommand{\thetable}{A.\arabic{table}}
\setcounter{table}{0} 

\startcontents[sections]
{	
	\hypersetup{linkcolor=black}
	\printcontents[sections]{l}{1}{}
}

\vspace{.3cm}

\noindent
This appendix provides implementation details (Section \ref{sec:app:implementation}), data analysis (Section \ref{sec:app:dataset-analysis}), additional
quantitative
(Section \ref{sec:app:additional-experiments})
and qualitative results (Section \ref{sec:app:qualitative-analysis}).
\new{We further refer to our project page for a supplementary video visualizing the results}.

\section{Implementation Details}
\label{sec:app:implementation}

This section provides additional implementation details for
LLM finetuning (\cref{sec:app:implementation:finetuning}),
prompt structure (\cref{sec:app:implementation:prompt_details}),
training data format (\cref{sec:app:implementation:training_data_format}),
\new{and the iterative prediction (\cref{sec:app:implementation:iterative_prediction}).}

\subsection{Finetuning the LLM}
\label{sec:app:implementation:finetuning}
As mentioned in \appendixref{Sec.~3}{\cref{sec:method}}, 
for all experiments,
we finetune Llama-3.1-8B-Instruct
model~\cite{dubey2024llama} using LoRA~\cite{hu2022lora}
with rank $r=8$ and target modules Q and V projections.
LoRA~\cite{hu2022lora} hyperparameters are set to $\alpha=32$ and $\text{dropout}=0.04$.
We use a batch size of 1 and a learning rate of $10^{-4}$,
and train for 1 epoch using the AdamW optimizer.
The training process takes 40 minutes using 4 NVIDIA H100 GPUs,
and inference on 100 short videos takes 30 minutes using the same hardware.

\subsection{Prompt details}
\label{sec:app:implementation:prompt_details}
The base prompt contains the instructions as follows:
\begin{lstlisting}[backgroundcolor = \color{backcolour}, basicstyle=\ttfamily]
 Given the complete transcript of a 
 video of duration {duration}, {task}.
 Identify the approximate start time 
 of each chapter in the format
 `hh:mm:ss - Title'.
 Ensure each chapter entry is on a new 
 line. 
 Focus on significant topic changes 
 that would merit a new chapter in a
 video, but do not provide summaries 
 of the chapters.
 {transcript}
\end{lstlisting}
\noindent
where \texttt{duration} represents the length of the video in \texttt{HH:MM:SS} format
(e.g., \texttt{00:09:52}), 
while \texttt{task} and \texttt{transcript} are specific to the input modalities used.

For example, when utilizing both ASR and captions as input modalities, the \texttt{task} is defined as follows:
\begin{lstlisting}[backgroundcolor = \color{backcolour}, basicstyle=\ttfamily,keywordstyle=\bfseries,morekeywords={captions, ASR}]
 use the provided captions and ASR 
 transcript to identify distinct 
 chapters based on content shifts.
\end{lstlisting}

\noindent
\new{For the \texttt{transcript},}
when training \method with both modalities, 
we prepend the modality names and interleave the outputs as illustrated below:
\begin{lstlisting}[backgroundcolor = \color{backcolour}, basicstyle=\ttfamily,
keywordstyle=\bfseries,morekeywords={ASR, Caption}]
 ASR 00:00:00: This place has blown 
   our minds.
 Caption 00:00:01: The image features 
   two individuals, a man and a woman, 
   standing outdoors in a natural 
   setting with rocky terrain and 
   sparse vegetation in the background.
 ASR 00:00:04: Look at this.
 ASR 00:00:05: In this episode, we're 
   exploring Buckhorn Wash, Utah.
\end{lstlisting}

\new{When training with only ASR (e.g., frame selector module), 
we simplify the input format by omitting the modality prefix,
as there is only one source of information in the transcript.}

We refer to \cref{tab:app:modality_prefix} for %
an experiment with/without these prefixes, \new{where we observe slight gains
by specifying the modalities}.
When using a single modality as input (e.g., ASR),
there is no need to prepend the modality name to the transcript:
\begin{lstlisting}[backgroundcolor = \color{backcolour}, basicstyle=\ttfamily]
 00:00:00: This place has blown 
   our minds.
 00:00:04: Look at this.
 00:00:05: In this episode, we're 
   exploring Buckhorn Wash, Utah.
\end{lstlisting}

\subsection{Training data format}
\label{sec:app:implementation:training_data_format}
For training our model, we use chapter data in the following structure. 
Each line contains the start timestamp of the chapter in \texttt{HH:MM:SS} format 
followed by the chapter title:

\begin{lstlisting}[backgroundcolor = \color{backcolour}, basicstyle=\ttfamily]
 00:00:00 - We're at Buckhorn Wash, 
   Utah
 00:00:51 - Morrison Knudson (MK) 
   Tunnels
 00:01:25 - In Buckhorn Wash, Like a 
   Little Zion
 00:02:15 - Buckhorn Wash Pictograph 
   Panel
 00:03:25 - Camping in the Wash, 
   Driving Through the Canyon
 00:04:47 - Swinging Bridge Campground 
   & San Rafael Bridge
 00:06:08 - Buckhorn Draw Visitor 
   Center, Well, & Spanish Trail
 00:08:37 - Boondocking at Utah Lake
 00:08:57 - Scenes from the Next 
   Episode - Nevada: Lemoille Canyon
 00:09:14 - Bloopers
\end{lstlisting}

\subsection{\new{Iterative prediction details}}
\label{sec:app:implementation:iterative_prediction}
\new{As mentioned in \appendixref{Sec.~3}{\cref{sec:method}}
and demonstrated through experiments in \appendixref{Sec.~4.4}{\cref{subsec:longer-videos}} of the main paper,
to handle videos with transcripts exceeding the LLM context window,
we implement an iterative prediction procedure using a sliding window approach. 
For each video, we segment the transcript into windows of fixed token length (e.g., 20k tokens) and process them sequentially. 
Starting from the first window, we generate chapters for the current segment, 
merge them with previously generated chapters, 
and advance the window to the next unprocessed portion of the transcript. 
This process continues until the entire video is covered.}

\section{Data Analysis and Statistics}
\label{sec:app:dataset-analysis}

Here, we provide a brief analysis %
of the portion
from the VidChapters dataset~\cite{yang2023vidchapters} that we used in our experiments.

\subsection{Video duration distribution}
\begin{figure*}
    \centering
    \includegraphics[width=\linewidth]{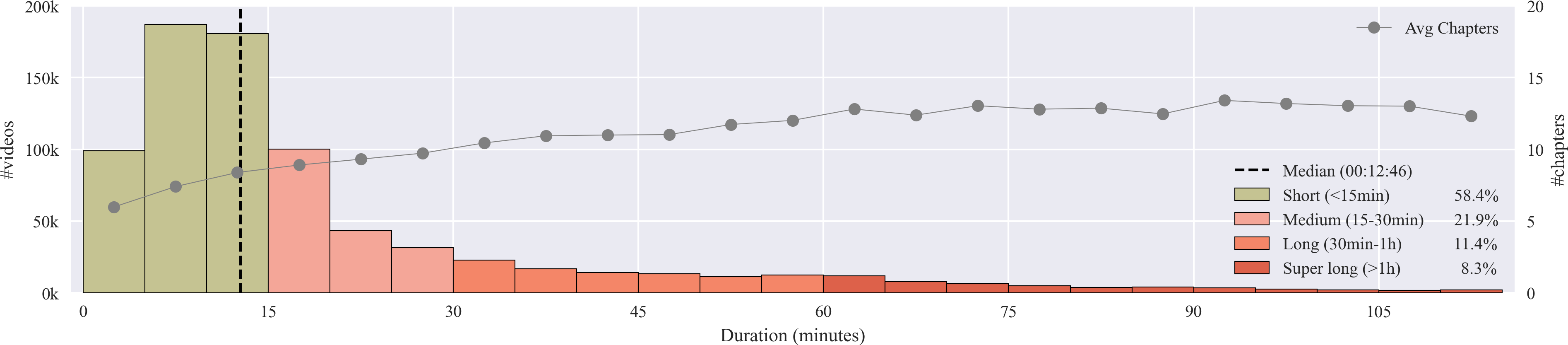}
    \vspace{-6mm}
    \caption{\textbf{Video duration distribution:}
    Distribution of video durations in our training set (bars, left axis) and average number of chapters per duration bin (gray line, right axis).
    Most videos are less than 15 minutes long, with progressively fewer videos at longer durations.
    The average number of chapters increases with video duration but plateaus around 13 chapters for videos longer than one hour.
    }
    \label{fig:app:video_durations}
\end{figure*}

Figure~\ref{fig:app:video_durations} 
shows the distribution of video durations in our training set.
The majority of videos (58.4\%) are short videos less than 15 minutes long,
while 21.9\% are medium-length (15-30 minutes), 
11.4\% are long (30-60 minutes), and 8.3\% exceed one hour.
Interestingly, 
we observe that the average number of chapters per video increases 
with video duration up to about 60 minutes, where it plateaus at approximately 13 chapters.
This plateau suggests a practical limit to manual chapter annotation, 
as annotators may be reluctant to segment videos into more than 13 chapters regardless of duration.
The median video duration is 12:46 minutes.

\subsection{\new{Video category distribution}}
\begin{figure}
    \centering
    \includegraphics[width=0.9\linewidth]{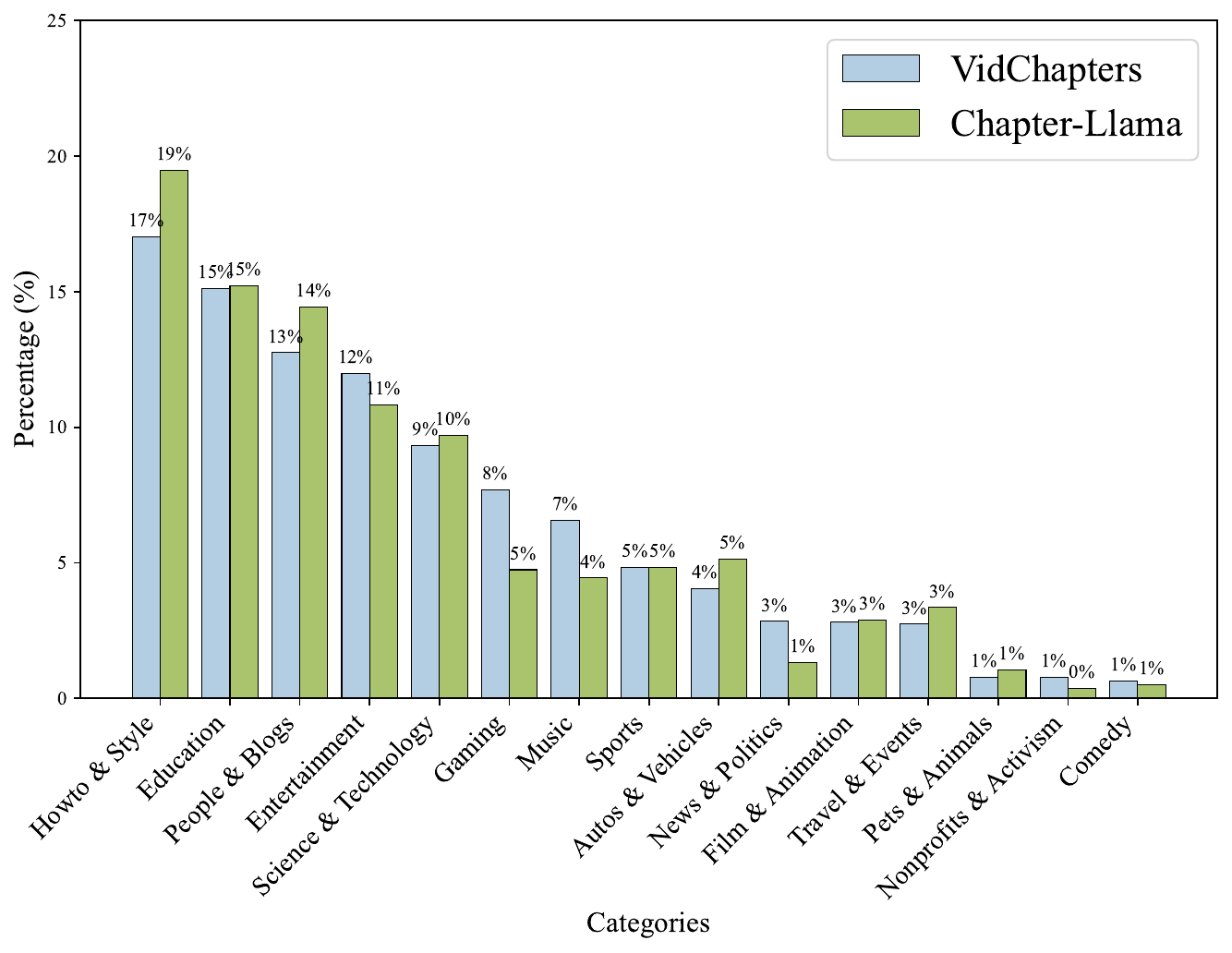}
    \caption{\textbf{\new{Video category distribution:}}
    \new{We compare the distribution of video categories
    	between the
        training set of the
        full
        VidChapters-7M dataset %
    	and
    	our 20k training subset.
    	We observe similar distributions given our uniform sampling from the original training set.}
    }
    \label{fig:app:video_categories}
\end{figure}

\new{For our final model, 
we use a subset of 20k training videos from VidChapters-7M.
Figure~\ref{fig:app:video_categories} compares the distribution of video categories 
between our training subset and the full VidChapters-7M dataset (Fig.~3~(d)~\cite{yang2023vidchapters}).
As we subsample uniformly from the original training set,
the two distributions closely match.}

\subsection{Videos within 15k window token limit}
Our models are trained with a context window of 15k tokens. 
In Table~\ref{tab:app:tokens_per_category},
we analyze the breakdown of videos across categories that 
fall within this limit.
All short and medium videos fall within this limit, while 79\% of long videos also comply.
Notably, for each category, 
the number of videos below the 15k token threshold exceeds 
the quantity required for model training before performance plateaus (see \appendixref{Fig.~4}{\cref{fig:datasize}} of the main paper).
This suggests that our current context window size 
is sufficient for effective training across all video duration categories.
Note we make this analysis with the full training set of the original VidChapters dataset, as our 20k subset considers videos that 100\% fall within the 15k limit.

\begin{table}
    \centering
    \setlength\tabcolsep{6pt}
    {
    \begin{tabular}{l|cc}   
    \toprule
    Category & \multicolumn{2}{c}{$<$15k tokens} \\
    \midrule
    Short & 466k  & 100 \% \\
    Medium & 175k & 100 \% \\
    Long & \z 71k & \z 79 \% \\
    \bottomrule
    \end{tabular}
    }
    \caption{\textbf{Videos in each category with fewer than 15k tokens:}
    We show the number of videos and proportion of short, medium, and long videos 
    in the training set that do not exceed the 15k token limit 
    of our training context window, from among 817k original training set videos of VidChapters.
    For videos without extracted captions, 
    the caption token length are estimated by 
    multiplying the average number of tokens per caption 
    by the number of ground truth chapters.}
    \label{tab:app:tokens_per_category}
\end{table}

\section{Additional Quantitative Results} %
\label{sec:app:additional-experiments}

\new{We report additional results
	with a range of experiments, such as the impact of input and output structure
	(\cref{subsec:app:nochaptertitles},~\ref{subsec:app:timestamp},~\ref{subsec:app:prefixes}),
	ablations with our frame selection,
	(\cref{subsec:app:alternative},~\ref{subsec:app:trainingsize},~\ref{subsec:app:separatetraining}),
	the LLM training,
	(\cref{subsec:app:llmvariants},~\ref{subsec:app:lorarank},~\ref{subsec:app:durations}),
	and further quantitative analyzes
	(\cref{subsec:app:oracle},~\ref{subsec:app:noasr},~\ref{subsec:app:metrics},~\ref{subsec:app:repetition},~\ref{subsec:app:numberofchapters}).}

\subsection{Predicting timestamps without chapter titles}
\label{subsec:app:nochaptertitles}
In our experiments, 
the \method model was trained to predict both chapter times and titles together. 
An alternative approach could involve training the model to predict chapter times exclusively, 
subsequently using another model to derive chapter titles from these times. 
However, as depicted in \cref{tab:app:predict_timestamps}, 
this approach underperforms compared to our current method. 
Therefore, we choose to continue training the \method model to predict both elements together, 
as the inclusion of chapter titles appears to enhance the accuracy of chapter time predictions.

\begin{table}
    \centering
    \setlength\tabcolsep{6pt}
    {
    \begin{tabular}{l|cccc}   
    \toprule
    Ground Truth Format & F1 & tIoU & S & C \\
    \midrule
    \texttt{HH:MM:SS} & 42.0 & 70.4  & - & - \\
    \texttt{HH:MM:SS - Title} & \textbf{42.6} & \textbf{70.6} & \textbf{16.4} & \textbf{82.4}\\
    \bottomrule
    \end{tabular}
    }
    \caption{\textbf{Effect of chapter titles on timestamp prediction:}
    We evaluate training \method with only timestamps or with timestamps and chapter titles,
    and observe that 
    adding chapter titles 
    slightly improves the segmentation metrics
    (F1: $+0.6$, tIoU: $+0.2$).
    }
    \label{tab:app:predict_timestamps}
\end{table}

\begin{table}
    \centering
    \setlength\tabcolsep{6pt}
    \resizebox{0.99\linewidth}{!}
    {
    \begin{tabular}{cc|l|cccc}   
    \toprule
    \multicolumn{2}{c|}{Modalities} & \multicolumn{1}{c|}{ASR} & \multicolumn{2}{c}{Segmentation} & \multicolumn{2}{c}{Titles}  \\
    Speech & Capt. &  timestamp & F1 & tIoU & S & C \\
    \midrule
    \multirow{2}{*}{\begin{tabular}[c]{@{}c@{}}{\cmark}\end{tabular}} 
    & \multirow{2}{*}{\begin{tabular}[c]{@{}c@{}}{-}\end{tabular}} 
        & start \, end & 41.4 & 69.7 & 15.8 & 77.9 \\
    &   & start & 38.5 & 68.1 & 13.9 & 67.3 \\
    \midrule
    \multirow{2}{*}{\begin{tabular}[c]{@{}c@{}}{\cmark}\end{tabular}} 
    & \multirow{2}{*}{\begin{tabular}[c]{@{}c@{}}{\cmark}\end{tabular}} 
        & start \, end & 39.1 & 67.6 & \z 6.0 & 19.9 \\
    &   & start & \textbf{42.6} & \textbf{70.6} & \textbf{16.4} & \textbf{82.4} \\
    \bottomrule
    \end{tabular}
    }
    \caption{\textbf{Adding end timestamps to ASR input:} 
    Adding end timestamps to ASR transcripts improves performance when using only speech (+2.9 F1).
    However, when combining speech with captions, including end timestamps decreases performance significantly, especially on title metrics (e.g., 19.9 vs 82.4 CIDEr). %
    We hypothesize this may be due to the inconsistency between modalities, where captions have single timestamps while speech segments have start and end times.
    }
    \label{tab:app:add_ASR_end}
\end{table}

\subsection{ASR timestamp representation}
\label{subsec:app:timestamp}
As mentioned in \appendixref{Sec.~3}{\cref{sec:method}},
we use ASR outputs obtained with WhisperX~\cite{bain2022whisperx},
which contain start and end timestamps of each ASR segment.
For our experiments, we only use the start timestamps,
as opposed to using start and end timestamps of each ASR segment.
In \cref{tab:app:add_ASR_end}, 
we analyze the impact of including end timestamps from ASR segments in addition to start timestamps. 
When using only speech inputs, including end timestamps improves performance (e.g., 41.4 vs 38.5 F1). %
However, when training with speech and captions,
using only start timestamps performs better, 
particularly for title generation metrics (e.g., 82.4 vs 19.9 CIDEr). %
We hypothesize this is because captions only have single timestamps, 
so having ASR segments with both start and end times creates an inconsistency between modalities that degrades performance. 
Therefore, in our final model we use only start timestamps for ASR segments.

\subsection{Modality prefixes}
\label{subsec:app:prefixes}
In \cref{tab:app:modality_prefix}, 
we analyze the impact of adding modality prefixes (``\texttt{ASR:}'' and ``\texttt{Caption:}'') 
before each text segment in the interleaved input sequence. 
Without prefixes, the model must infer the modality type implicitly - for captions this may be easier since they %
often
start with 
``The image shows'', while ASR segments have varied structure. %
Results show that explicitly marking modalities with prefixes improves 
performance across all metrics (e.g., 42.6 vs 41.9 F1), %
suggesting that helping the model distinguish between modalities is beneficial.

\begin{table}
    \centering
    \setlength\tabcolsep{6pt}
    {
    \begin{tabular}{c|cccc}   
    \toprule
    Has prefix? & F1 & tIoU & S & C \\
    \midrule
    \xmark & 41.9 & 69.6 & 16.0 & 78.5 \\
    \cmark  & \textbf{42.6} & \textbf{70.6} & \textbf{16.4} & \textbf{82.4} \\
    \bottomrule
    \end{tabular}
    }
    \caption{\textbf{Effect of modality prefixes:} 
    Adding prefixes to the ASR and captions modalities improves performance.
    }
    \label{tab:app:modality_prefix}
\end{table}

\begin{table}
    \centering
    \resizebox{0.99\linewidth}{!}
    {
    \begin{tabular}{lc|cccc}   
    \toprule
    Frame selection for captions & \#frames $\downarrow$ & F1 & tIoU & S & C \\
    \midrule
    Shot midpoints & 49.4
        & 40.8 & 69.1 & 15.6 & 77.0 \\
    Shot boundaries & 49.4
        & 40.6 & 69.1 & 15.8 & 79.3 \\
    Speech-based CL $\pm$1 sec & 20.6 
        & \textbf{42.7} & 69.5 & \textbf{16.5} & \textbf{83.2} \\
    Speech-based CL midpoints & \textbf{10.3} 
        & 41.2 & 69.0 & 15.6 & 73.7 \\
    Speech-based CL boundaries & \textbf{10.3} 
        & 42.6 & \textbf{70.6} & 16.4 & 82.4  \\
    \bottomrule
    \end{tabular}
    }
    \caption{\textbf{\new{Alternative frame selection strategies:}}
    \new{We evaluate alternative frame sampling strategies including:
    (1) shot boundaries and midpoints detected with \texttt{PySceneDetect}~\cite{brandon2018},
    (2) frames sampled $\pm$1 second around chapter boundaries predicted by our speech-based Chapter-Llama (CL) model,
    (3) frames at CL predicted boundaries and midpoints between them.
    Results show that sampling at CL boundaries achieves competitive performance across all metrics 
    while requiring significantly fewer frames (10.3 vs 20.6-49.4 frames per video).
    }
    }
    \label{tab:app:frame-selector_alternatives}
\end{table}

\subsection{\new{Alternative frame selection strategies}}
\label{subsec:app:alternative}
\new{In the main paper,
given a detected chapter boundary from our speech-only model, 
we select frames at the boundary location itself. 
In \cref{tab:app:frame-selector_alternatives}, we explore alternative frame sampling strategies,
including:
(1)~shot boundaries or midpoints detected with \texttt{PySceneDetect}~\cite{brandon2018},
(2)~$\pm1$~sec before and after speech-based chapter boundary predictions,
(3)~speech-based Chapter-Llama (CL) predicted boundary locations and midpoints between these locations.
See the caption for comments.}

\subsection{Training data size on the frame selection model}
\label{subsec:app:trainingsize}
Throughout our experiments,
we train the speech-only model using
10k videos 
to obtain frame locations for caption extraction
(and 1k videos in most of our experiments to train our \method model).
In \cref{tab:app:data_size_asr_preds},
we analyze how the amount of training data in the speech-only model affects 
downstream performance on our \method model using both speech and captions.

The second to last row (42.6 F1) represents our main result reported in our ablations, 
and the last row (46.7 F1)
shows results when using 10k videos for speech-only model training 
and 10k videos for \method (CL) model training, corresponding to the
final point in the \textit{number of training videos vs performance} plot in \appendixref{Fig.~4}{\cref{fig:datasize}} of the main paper. 
The first two rows show new results using only 1k videos to train the speech-only model. 
We observe that increasing training data for the speech-only frame selector model from 1k to 10k videos
has minimal impact on segmentation metrics but improves title generation performance in both cases -- 
from 17.5 to 18.6 SODA when using 10k videos for \method training, 
and from 15.6 to 16.4 SODA when using 1k videos for \method training. 
Increasing the training data from 1k to 10k videos for our \method model
improves performance on both segmentation and title benchmarks,
with F1 scores improving from 42.7 to 46.9 and from 42.6 to 46.7, respectively.

\begin{table}
    \centering
    \setlength\tabcolsep{6pt}
    {
    \begin{tabular}{cc|cccc}   
    \toprule
    \multicolumn{2}{c|}{\# videos} & \multicolumn{2}{c}{Segmentation} & \multicolumn{2}{c}{Titles}  \\
    F. selector & CL & F1 & tIoU & S & C \\
    \midrule
    \multirow{2}{*}{\begin{tabular}[c]{@{}c@{}}{\z 1k}\end{tabular}} 
    & \z 1k   & 42.7 & 70.8 & 15.6 & 78.1 \\
    & 10k     & \textbf{46.9} & \textbf{72.9} & 17.5 & 86.8\\
    \midrule
    \multirow{2}{*}{\begin{tabular}[c]{@{}c@{}}{10k}\end{tabular}} 
    & \z 1k     & 42.6 & 70.6 & 16.4 & 82.4 \\
    & 10k       & 46.7 & 72.2 & \textbf{18.6} & \textbf{96.4} \\
    \bottomrule
    \end{tabular}
    }
    \caption{\textbf{Effect of training data size on speech-based frame selector:}
    We analyze how the amount of training data used for the speech-only frame selector (first column) affects downstream performance of our \method (CL) model.
    The frame selector is trained on either 1k or 10k videos to predict frame locations where captions should be extracted, 
    while the CL is trained on either 1k or 10k different videos for chapter generation.
    \new{Comparing rows 1 vs 3 and 2 vs 4, 
    we observe that increasing frame selector training data from 1k to 10k videos 
    has minimal impact on segmentation metrics,
    but slightly improves title generation.
    In contrast, 
    increasing CL training data from 1k to 10k videos 
    (rows 1 vs 2 and 3 vs 4)
    improves both segmentation and title metrics.}
    }
    \label{tab:app:data_size_asr_preds}
\end{table}

\subsection{Separate training data for frame selector and \method}
\label{subsec:app:separatetraining}
In all our experiments, 
we use a different subset of videos to train the frame selector model and the \method model.
In \cref{tab:app:frame-selector_data-overlap},
we analyze the performance of \method when using the same set of 1k videos for both models 
or when using a different set of 1k videos for the \method model.
We see that using the same set of videos for both models decreases performance.
We hypothesize that this performance drop occurs due to overfitting in the training pipeline:
When both models are trained on the same videos, 
the outputs of the frame selector
align very closely with the ground truth locations for those specific videos.
This creates an artificial correlation between frame locations and content that the \method model learns to exploit during training.
As a result, \method develops an over-reliance on the precise temporal positions of frames rather than learning to
refine the
location %
information.

 \begin{table}
    \centering
    \setlength\tabcolsep{6pt}
    {
    \begin{tabular}{c|cccc}   
    \toprule
    Training data & F1 & tIoU & S & C \\
    \midrule
    $V_{F.S.} = V_{C.L.}$    & 41.4 & 70.1 & 15.1 & 77.5 \\
    $V_{F.S.} \neq V_{C.L.}$ & \textbf{42.7} & \textbf{70.8} & \textbf{15.6} & \textbf{78.1} \\
    \bottomrule
    \end{tabular}
    }
    \caption{\textbf{Frame selector and \method training data overlap:} 
    Given the set of videos used to train the speech-based frame selector model ($V_{F.S.}$) and 
    and
    the \method model ($V_{C.L.}$),
    we compare the performance of \method when using 
    different subsets of videos ($V_{F.S.} \neq V_{C.L.}$), and
    when
    using the same, already seen, videos ($V_{F.S.} = V_{C.L.}$).
    We see that using the same 1k set of videos for both models decreases performance.
    }
    \label{tab:app:frame-selector_data-overlap}
\end{table}

\subsection{\new{LLM variants}}
\label{subsec:app:llmvariants}
\new{%
We conduct experiments with different variants of the Llama model family.
All our previous results use Llama-3.1-8B-Instruct,
and
we now compare it against the more recent Llama-3.2 model in three sizes: 1B, 3B, and 11B parameters.

As shown in \cref{tab:app:llama_variants},
model size has a significant effect on chaptering quality.
Using speech only, 
the F1 score improves substantially from 23.5 to 35.2 to 38.5 as we scale from 1B to 3B to 8B parameters, with only a minor additional gain to 39.8 when scaling to 11B parameters.
This trend holds across all metrics.
Llama-3.1-8B performs similar to Llama-3.2-11B,
which we use in our final model due to reduced computational complexity.
Note that we were unable to run Llama-3.2-11B on our final
model combining speech and captions due to hardware constraints.
}

\begin{table}
    \centering
    \resizebox{0.99\linewidth}{!}
    {
    \begin{tabular}{c|cc|cc|ccc}   
    \toprule
    Llama & Speech & Captions & F1 & tIoU & S & C \\
    \midrule
    \multirow{2}{*}{Llama-3.2-1B} 
        & \cmark & -        & 23.5 & 58.3 & \z 6.9 & 23.9  \\
        & \cmark & \cmark   & 24.6 & 58.6 & \z 7.4 & 28.0 \\
    \midrule
    \multirow{2}{*}{Llama-3.2-3B} 
        & \cmark & -        & 35.2 & 66.7 & 10.5 & 52.5 \\
        & \cmark & \cmark   & 34.7 & 65.2 & 12.5 & 63.6 \\
    \midrule
    \multirow{2}{*}{Llama-3.2-11B} 
        & \cmark & -        & 39.8 & 67.9 & 14.8 & 71.1 \\
        & \cmark & \cmark   & n/a & n/a & n/a & n/a \\
    \midrule
    \multirow{2}{*}{Llama-3.1-8B} 
        & \cmark & -        & 38.5 & 68.1 & 13.9 & 67.3 \\
        & \cmark & \cmark   & \textbf{42.6} & \textbf{70.6} & \textbf{16.4} & \textbf{82.4} \\
    \bottomrule
    \end{tabular}
    }
    \caption{\textbf{\new{Llama variants:}}
    \new{ 
    Model size has a significant impact on performance on Llama3.2 family. Llama-3.1-8B remains our choice due to its competitive performance with manageable computational complexity.
    }
    }
    \label{tab:app:llama_variants}
\end{table}

\subsection{LoRA rank}
\label{subsec:app:lorarank}
In \cref{tab:app:lora_rank}, we conduct experiments comparing LoRA ranks $r=8$ and $r=16$ across different training data sizes.
With 1k training videos, the lower rank $r=8$ performs notably better
(42.6 vs 39.9 F1 score).
As we increase to 5k videos, $r=16$ shows a slight advantage
(46.5 vs 45.6 F1),
while at 10k videos both ranks achieve comparable performance
(46.7 vs 46.6 F1).
This suggests that with limited training data,
a lower rank helps prevent overfitting,
while with more data the model capacity becomes less critical.
Based on these findings and considering
efficiency,
we use $r=8$ as our default LoRA rank throughout all experiments in the paper.

\begin{table}
    \centering
    \setlength\tabcolsep{6pt}
    {
    \begin{tabular}{cc|cccc}   
    \toprule
    \#videos & rank & F1 & tIoU & S & C \\
    \midrule
    \multirow{2}{*}{\begin{tabular}[c]{@{}c@{}}{\z 1k}\end{tabular}} 
    & \z 8  & 42.6 & 70.6 & 16.4 & 82.4 \\
    & 16    & 39.9 & 68.5 & 15.6 & 78.4 \\
    \midrule
    \multirow{2}{*}{\begin{tabular}[c]{@{}c@{}}{\z 5k}\end{tabular}} 
    & \z 8  & 45.6 & 72.3 & 18.3 & 90.0 \\
    & 16    & 46.5 & 72.8 & 18.5 & 92.8 \\
    \midrule
    \multirow{2}{*}{\begin{tabular}[c]{@{}c@{}}{10k}\end{tabular}} 
    & \z 8  & 46.7 & 72.2 & 18.6 & 96.4 \\
    & 16    & 46.6 & 72.4 & 18.6 & 92.5 \\
    \bottomrule
    \end{tabular}
    }
    \caption{\textbf{\new{LoRA rank:}}
    \new{Comparing LoRA ranks r=8 and r=16, 
    we find that with 1k training videos, 
    the lower rank performs better. 
    With 5k videos, 
    r=16 slightly outperforms r=8.
    At 10k videos, 
    both ranks achieve similar results, 
    suggesting that
    with sufficient training data, 
    model capacity becomes less important.
    }
    }
    \label{tab:app:lora_rank}
\end{table}

\begin{table*}
\centering
\resizebox{0.99\linewidth}{!}
{
\begin{tabular}{@{}l|cccc|cccc|cccc|cccc@{}}
\toprule
    Training & \multicolumn{4}{c|}{Short (val)} & \multicolumn{4}{c|}{Medium (val)} & \multicolumn{4}{c|}{Long (val)} & \multicolumn{4}{c}{All (val)} \\
    videos & F1 & tIoU & S & C & F1 & tIoU  & S & C & F1 & tIoU  & S & C & F1 & tIoU  & S & C \\
    \midrule
    Short 
        & \textbf{49.7} & \textbf{75.0} & \textbf{21.4} & \textbf{112.9}
        & 38.3 & 67.6 & {13.2} & {61.4}
        & 37.9 & 66.7 & 12.8 & 63.3
        & 42.0 & 69.8 & 15.8 & 79.2 \\
    \new{Medium} 
        & 47.5 & 74.6 & 21.3 & 109.8
        & 37.9 & 67.5 & {13.2} & 55.6
        & 38.3 & 67.0 & 13.3 & 63.5
        & 41.2 & 69.7 & 15.9 & 76.3\\
    \new{Long} 
        & 46.6 & 74.0 & 19.5 & 104.9
        & \textbf{39.3} & \textbf{68.1} & \textbf{13.4} & \textbf{62.0}
        & 38.1 & 66.9 & 14.3 & 75.1
        & 41.3 & 69.7 & 15.8 & 80.8\\
    All 
        & 48.4 & 74.4 & 21.2 & 110.8
        & {38.9} & {68.0} & 13.1 & 57.3
        & \textbf{40.4} & \textbf{69.3} & \textbf{14.9} & \textbf{79.1}
        & \textbf{42.6} & \textbf{70.6} & \textbf{16.4} & \textbf{82.4} \\
\bottomrule
\end{tabular}
}
\vspace{-0.2cm}
\caption{\textbf{Including long videos at training improves results:} 
Training with \new{1k videos balanced across short,
medium, and long durations} (last row, `All') 
improves performance compared to 
training with just 1k short videos (first row).
\new{The improvement is most pronounced for long videos (+2.5 F1)}.
When averaging across short/medium/long validation splits, training with all videos improves all metrics: 
\new{F1 (+0.6), tIoU (+0.8), S (+0.6), and C (+3.2)}.
}
\vspace{-0.2cm}
\label{tab:app:long_videos}
\end{table*}

\subsection{Training on videos of various durations}
\label{subsec:app:durations}
\new{In most of our experiments, %
we have trained our model on 1k videos balanced across duration categories,
i.e.,
333 short videos ($<$15 min), 
333 medium-length videos (15-30 min),
and 334 long videos (30-60 min).
In \cref{tab:app:long_videos}, 
we show the benefit of such training on videos of various durations.
For this experiment, we train new models
only on 1k short videos, 
on 1k medium videos, and
on 1k long videos. %
For evaluation, we use the same 300 validation videos as before, 
with 100 videos sampled from each duration category.
As expected, 
training on short videos performs best on short videos (49.7 F1),
while training on long videos performs best on long videos (40.4 F1). 
Training with a balanced mix of all three durations achieves the best overall performance across all categories (42.6 F1).}

\subsection{~Oracle experiments with partial ground truth input}
\label{subsec:app:oracle}
To evaluate the \method model's capability in predicting 
chapters when provided with ground-truth chapter boundaries or titles, 
we conduct experiments with two scenarios: 
(i)~incorporating ground truth timestamps into the input, and 
(ii)~including ground truth chapter titles. 
In the first scenario, the task represents an upper bound limit 
of title metrics
for our model, as it predicts chapters based on known timestamps. 
In the second scenario, the model predicts chapters using known titles, 
serving as a form of video chapter grounding. 
As demonstrated in \cref{tab:app:given_gt}, 
these experiments establish the upper bounds of our model's performance.
\begin{table}
    \centering
    \setlength\tabcolsep{6pt}
    {
    \begin{tabular}{cc|cccc}   
    \toprule
    Boundaries & Titles & F1 & tIoU & S & C \\
    \midrule
    \xmark & \xmark & 42.6 & 70.6 & 16.4 & \z 82.4 \\
    \gray{\cmark} & \gray{\xmark} & \gray{99.1} & \gray{99.7} & \gray{23.8} & \gray{121.4} \\
    \gray{\xmark} & \gray{\cmark} & \gray{64.0} & \gray{80.1} & \gray{71.5} & \gray{506.3} \\
    \bottomrule
    \end{tabular}
    }
    \caption{\textbf{Oracle experiment with partial ground truth input:} 
    We evaluate the capability of \method in predicting chapters 
    when provided with ground truth chapter boundaries or titles. 
    The first scenario represents an oracle experiment for title metrics,
    as it predicts chapters based on known timestamps (second row).
    The second scenario serves as a form of video chapter grounding, i.e.,
    given known titles to segment the boundaries (last row).
    The model was trained with 1k videos and evaluated with \new{300} videos.}
    \label{tab:app:given_gt}
\end{table}

\subsection{~Performance on videos that have no speech}
\label{subsec:app:noasr}
As mentioned in \appendixref{Sec.~4}{\cref{sec:experiments}},
most of the videos ($>97\%$) in the dataset have speech content.
For the videos that have no ASR detections, we use every 10s sampling.
We now investigate the performance of our approach when there is no ASR available. 
In~\cref{tab:app:no_asr_val}, 
\new{we select all videos in the validation set
without ASR, totaling 190 videos,}
and compare the performance to Vid2Seq~\cite{yang2023vid2seq}.
We observe that the performance of both models is
worse than when ASR is available,
suggesting that both models mainly benefit from speech input.
However, our approach still outperforms Vid2Seq in this challenging setting.
By visually inspecting some of these videos, we noticed
failure cases with music videos,
with very similar backgrounds across frames, which makes it difficult for the model to detect chapter boundaries without any audio information.
This is left to future work, as stated in the conclusions of the main paper.
We also notice success cases often depict frames with text,
which are captured by the captioner (see first and last examples in \cref{fig:app:qualitative_no-asr}).

\begin{table}
    \centering
    \setlength\tabcolsep{6pt}
    {
    \begin{tabular}{l|cccc}   
    \toprule
    Method & F1 & tIoU & S & C \\
    \midrule
    Vid2Seq~\cite{yang2023vid2seq} & 12.6 & 45.5 & 5.5 & 18.0 \\
    \method (ours) & \textbf{15.5} & \textbf{49.6} & \textbf{5.0} & \textbf{26.3}  \\
    \bottomrule
    \end{tabular}
    }
    \caption{\textbf{Performance on validation videos without ASR:} 
    We evaluate the performance of our best performing model in videos without ASR predictions 
    (190 videos in validation).
    We observe that the \method outperforms Vid2Seq in all metrics, 
    but the performance of both models is worse than when ASR is available.
    }
    \label{tab:app:no_asr_val}
\end{table}

\subsection{~Full set of metrics} %
\label{subsec:app:metrics}
In \appendixref{Sec.~4.1}{\cref{subsec:data-evaluation}} of the main paper, 
we adopted the evaluation metrics (F1, tIoU, SODA, and CIDEr),
which we consider more suitable for assessing video chapter generation.
For completeness and direct comparison with VidChapters~\cite{yang2023vidchapters},
we also report results using their full set of metrics 
in \cref{tab:app:metrics_global,tab:app:metrics_segmentation}.
The segmentation metrics include precision and recall at 3-second and 5-second thresholds,
as well as at 0.5 and 0.7 IoU thresholds.
The full metrics (referred to as `global metrics' by \cite{yang2023vidchapters}) comprise 
SODA (S)~\cite{fujita2020soda}, 
BLEU (B1-B4)~\cite{papineni2002bleu}, 
CIDEr (C)~\cite{vedantam2015cider}, 
METEOR (M)~\cite{banerjee2005meteor}, 
and ROUGE-L (RL)~\cite{lin2004rouge}.
Our model consistently outperforms Vid2Seq~\cite{yang2023vid2seq} across all metrics.

\begin{table*}
    \centering
    \setlength\tabcolsep{6pt}
    {
    \begin{tabular}{l|cc|cc|cc|cc}   
    \toprule
    Method & P@5s & R@5s & P@3s & R@3s & P@0.5 & R@0.5 & P@0.7 & R@0.7\\
    \midrule
    Vid2Seq~\cite{yang2023vid2seq} & 30.6 & 36.4 & 24.4 & 28.7 & 46.3 & 51.1 & 28.7 & 30.6 \\
    \method & \textbf{52.0} & \textbf{51.7} & \textbf{45.1} & \textbf{44.7} & \textbf{66.3} & \textbf{63.4} & \textbf{49.9} & \textbf{47.8} \\
    \bottomrule
    \end{tabular}
    }
    \caption{\textbf{Video chapter generation (segmentation metrics) on VidChapters~\cite{yang2023vidchapters} test set:} 
    Comparison of segmentation metrics between Vid2Seq and our best model from \appendixref{Tab.~1}{\cref{tab:sota}}.
    Metrics include 
    precision and recall at 3-second and 5-second thresholds,
    as well as at 0.5 and 0.7 IoU thresholds.    
    Our method consistently outperforms Vid2Seq across all metrics.
    }
    \label{tab:app:metrics_segmentation}
\end{table*}

\begin{table}
    \centering
    \setlength\tabcolsep{6pt}
    \resizebox{0.99\linewidth}{!}
    {
    \begin{tabular}{l|cccccccc}   
    \toprule
    Method & S & B1 & B2 & B3 & B4 & C & M & RL\\
    \midrule
    Vid2Seq~\cite{yang2023vid2seq} & 11.6 & 11.1 & \z 7.7 & 4.5 & 3.1 & \z 55.8 & \z 9.6 & 12.8 \\
    \method & \textbf{19.3} & \textbf{19.5} & \textbf{14.3} & \textbf{8.7} & \textbf{5.6} & \textbf{100.9} & \textbf{15.4} & \textbf{22.2} \\
    \bottomrule
    \end{tabular}
    }
    \caption{\textbf{Full metrics used by VidChapters~\cite{yang2023vidchapters}:} 
    We report the full metrics (referred to as `global metrics' in \cite{yang2023vidchapters}) on
    the test set of VidChapters.
    We compare Vid2Seq and
    our best model from \appendixref{Tab.~1}{\cref{tab:sota}}.
    Metrics include 
    SODA~\cite{fujita2020soda} (S), 
    BLEU~\cite{papineni2002bleu} (B1-B4), 
    CIDEr~\cite{vedantam2015cider} (C), 
    METEOR~\cite{banerjee2005meteor} (M), 
    and ROUGE-L~\cite{lin2004rouge} (RL).
    Our method consistently outperforms Vid2Seq across all metrics.
    }
    \label{tab:app:metrics_global}
\end{table}

\subsection{~Repetition analysis}
\label{subsec:app:repetition}
We have noticed that Vid2Seq tends to repeat chapter titles 
(see \appendixref{Fig.~3}{\cref{fig:qualitative}} of the main paper).
To quantify this, 
we calculate the ratio of unique chapter titles to the total 
number of chapter titles predicted for each video
and then average this ratio across all videos in the test set.
For the ground truth, this average ratio is 99.6\%, i.e., almost all chapter titles are unique.
For our finetuned model, this average ratio is 96.3\%.
In contrast, Vid2Seq has a much lower average ratio of 63.5\%, 
indicating that it indeed repeats chapter titles frequently.

\subsection{~Accuracy of number of chapter predictions}
\label{subsec:app:numberofchapters}
\begin{figure}
\centering
\includegraphics[width=\linewidth]{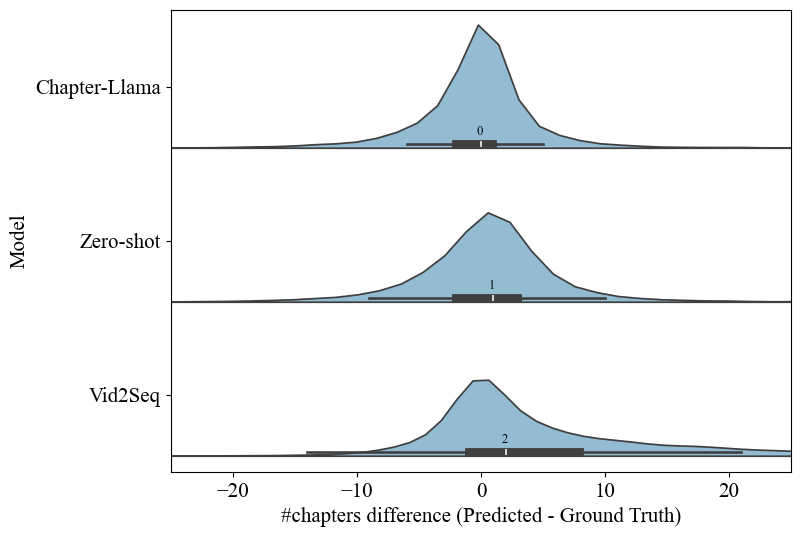}
\vspace{-6mm}
\caption{\textbf{Accuracy of number of chapter predictions:}
	The violin plot shows the distribution of differences between 
	the predicted and ground truth number of chapters for three video chaptering models:
	\method, Zero-shot, and Vid2Seq.
	The \method model exhibits the most concentrated distribution centered around 0, 
	indicating accurate number of chapter prediction.
	The Zero-shot model tends to slightly overpredict the number of chapters,
	while the Vid2Seq model often significantly overpredicts the number of chapters. 
	The median differences are 0, 1, and 2 for Chapter-Llama, Zero-shot, and Vid2Seq, respectively, with mean number of chapter differences of -0.2, 0.5, and 4.5 (not shown). 
}
\label{fig:app:number_of_chapters}
\end{figure}

While our main evaluation focused on the quality of chapter segment predictions,
it is also important to assess the accuracy in predicting the number of chapters.
Our primary metrics (F1, tIoU, SODA, and CIDEr) do not directly indicate whether
the predicted chapter count is correct or if the method tends to over- or under-segment.
To evaluate this, we analyze the distribution of differences between predicted and
ground truth chapter counts for \method, Zero-shot, and Vid2Seq models,
as illustrated in \cref{fig:app:number_of_chapters}.

The results reveal that \method exhibits the most concentrated distribution
centered around zero, indicating superior accuracy in predicting chapter counts.
In contrast, both Zero-shot and Vid2Seq models over-segments the video with a high number of chapters. 
The tight interquartile range and symmetrical density shape of \method suggest
a more
reliable %
chapter count prediction.
However, it is important to note that accurately predicting the number of chapters
does not necessarily guarantee correct chapter segmentation.

\section{Additional Qualitative Analyses}
\label{sec:app:qualitative-analysis}

\new{We present %
	several qualitative analyses:
(i)~evaluation metric calculation examples (\cref{subsec:app:evaluation_metrics}),
(ii)~caption visualizations %
(\cref{subsec:app:caption_examples}),
and
(iii)~predictions from our model (\cref{subsec:app:qualitative_examples}).}

\subsection{Evaluation metrics}
\label{subsec:app:evaluation_metrics}
\begin{figure*}
	\centering
	\includegraphics[width=.99\linewidth]{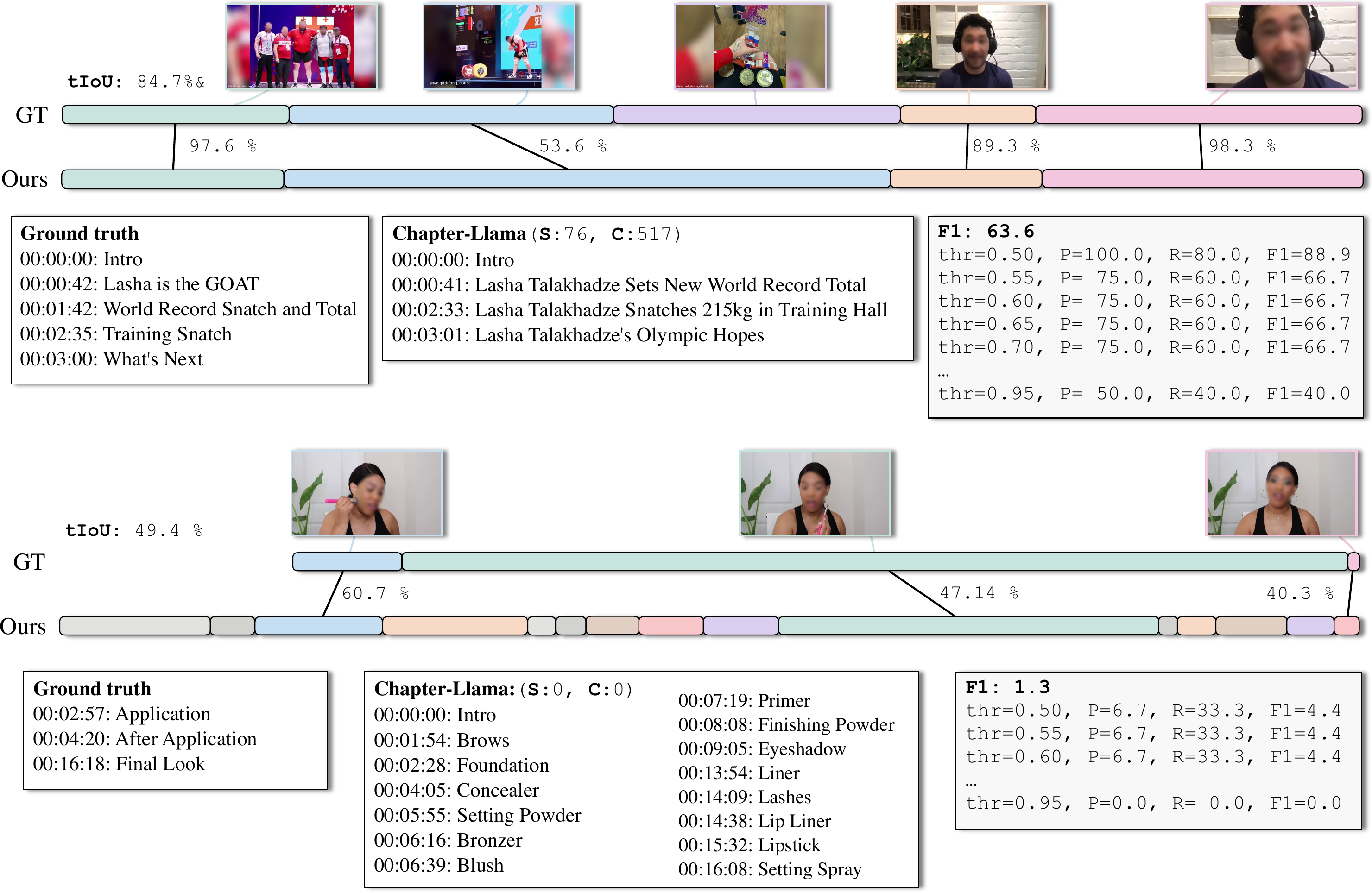}
    \vspace{-0.3cm}
	\caption{\textbf{Segmentation metrics visualization:} 
	We illustrate with examples how tIoU and F1 scores are calculated for video chaptering. 
	The top example shows a high-quality prediction with good overlap, 
	while the bottom example demonstrates a lower-quality prediction with more misalignments.
    We additionally show the corresponding SODA (S) and CIDEr (C) scores.
	}
	\label{fig:app:segmentation_metrics}
\end{figure*}

In \appendixref{Sec.~4.1}{\cref{subsec:data-evaluation}},
we introduced our primary evaluation metrics for video chaptering: \textbf{tIoU} and \textbf{F1} scores.
Here, we illustrate how these metrics are calculated using concrete examples, as shown in \cref{fig:app:segmentation_metrics}.

For tIoU (temporal Intersection over Union), 
we first match predicted and ground truth segments by greedily selecting pairs with the highest IoU scores. 
In the top example of \cref{fig:app:segmentation_metrics}, 
we have 5 ground truth chapters and 4 predicted chapters.
The matching process starts with chapters having the most overlap, 
and each chapter can only be used once. 
The tIoU score (84.7) is then calculated as the mean IoU across all matched pairs 
(97.6, 53.6, 89.3, 98.3). 
Similarly, for the bottom example, the tIoU score of 49.4 is the mean of 60.7, 47.14, and 40.3.

For the F1 score, we compute precision and recall at different IoU thresholds 
(from 0.5 to 0.95 with a step of 0.05). 
In the top example, at a threshold of 0.5,
all predicted chapters have a ground truth match with an overlap higher than 50\%, 
resulting in a precision of 100\%. 
However, one ground truth chapter out of 5 is left without a prediction, 
leading to a recall of 80\%. 
The F1 score is then computed as the harmonic mean of precision and recall. 
This process is repeated for all thresholds, 
and the final F1 metric is the average across these thresholds.

\subsection{Visualizing captions}
\label{subsec:app:caption_examples}
\begin{figure*}
	\centering
	\includegraphics[width=.99\linewidth]{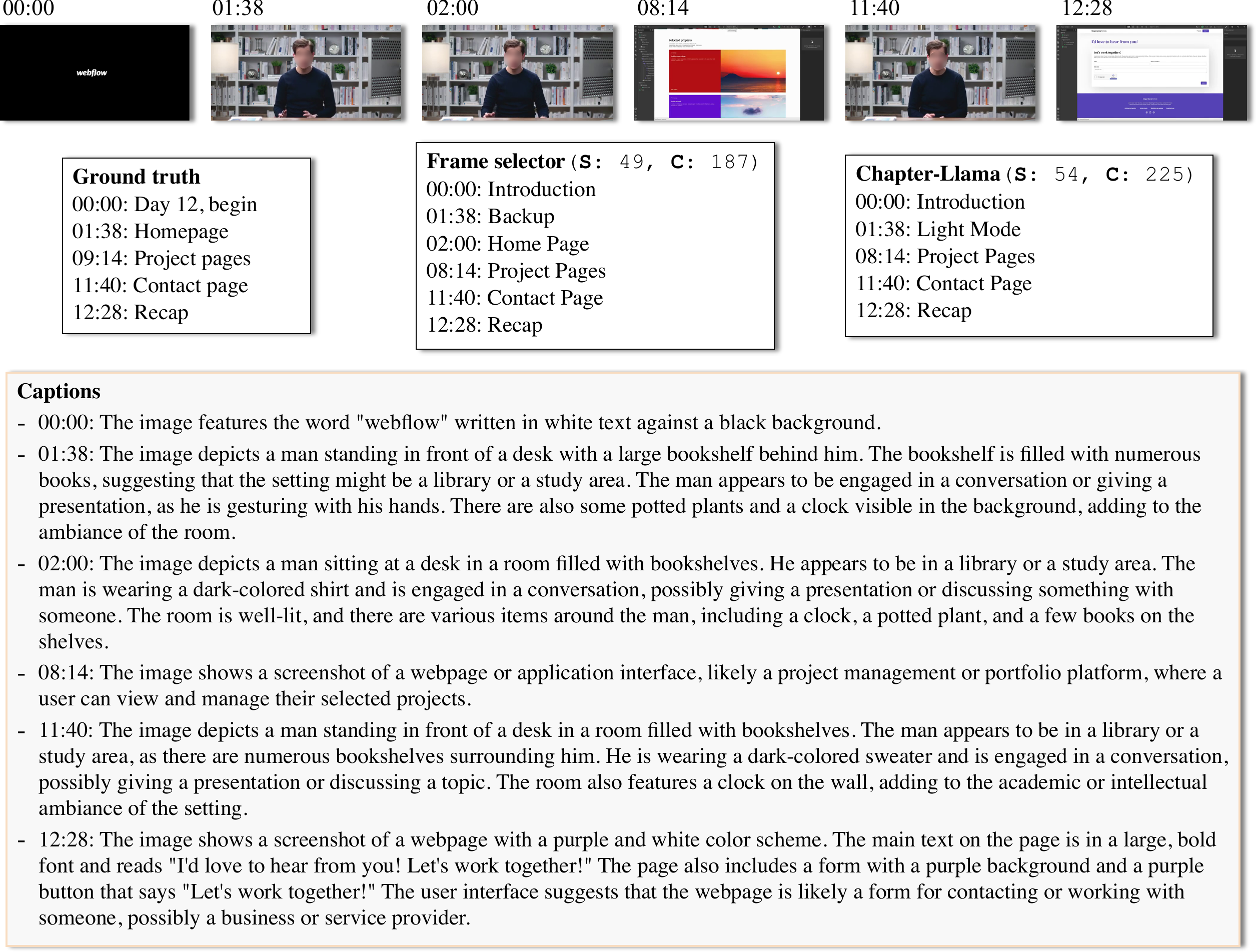}
	\caption{\textbf{Visualizing captions:} 
	We provide an example with chapter predictions using the speech-based frame selection model,
    the corresponding captions sampled,
    and the refined predictions produced by \method.
    We additionally show the corresponding SODA (S) and CIDEr (C) scores.
    We see that the initially predicted chapter at timestamp 02:00 is suppressed by \method.
	}
	\label{fig:app:caption_frame_selection}
\end{figure*}

In \cref{fig:app:caption_frame_selection},
we provide an example, where we also visualize some of the
intermediate captions that are fed to our chapter generation LLM.
We then show the chapter predictions from the
speech-based frame selection model,
the corresponding captions selected based on this model,
and the refined predictions with \method.

\subsection{\method prediction examples}
\label{subsec:app:qualitative_examples}
\begin{figure*}
	\centering
	\includegraphics[width=.99\linewidth]{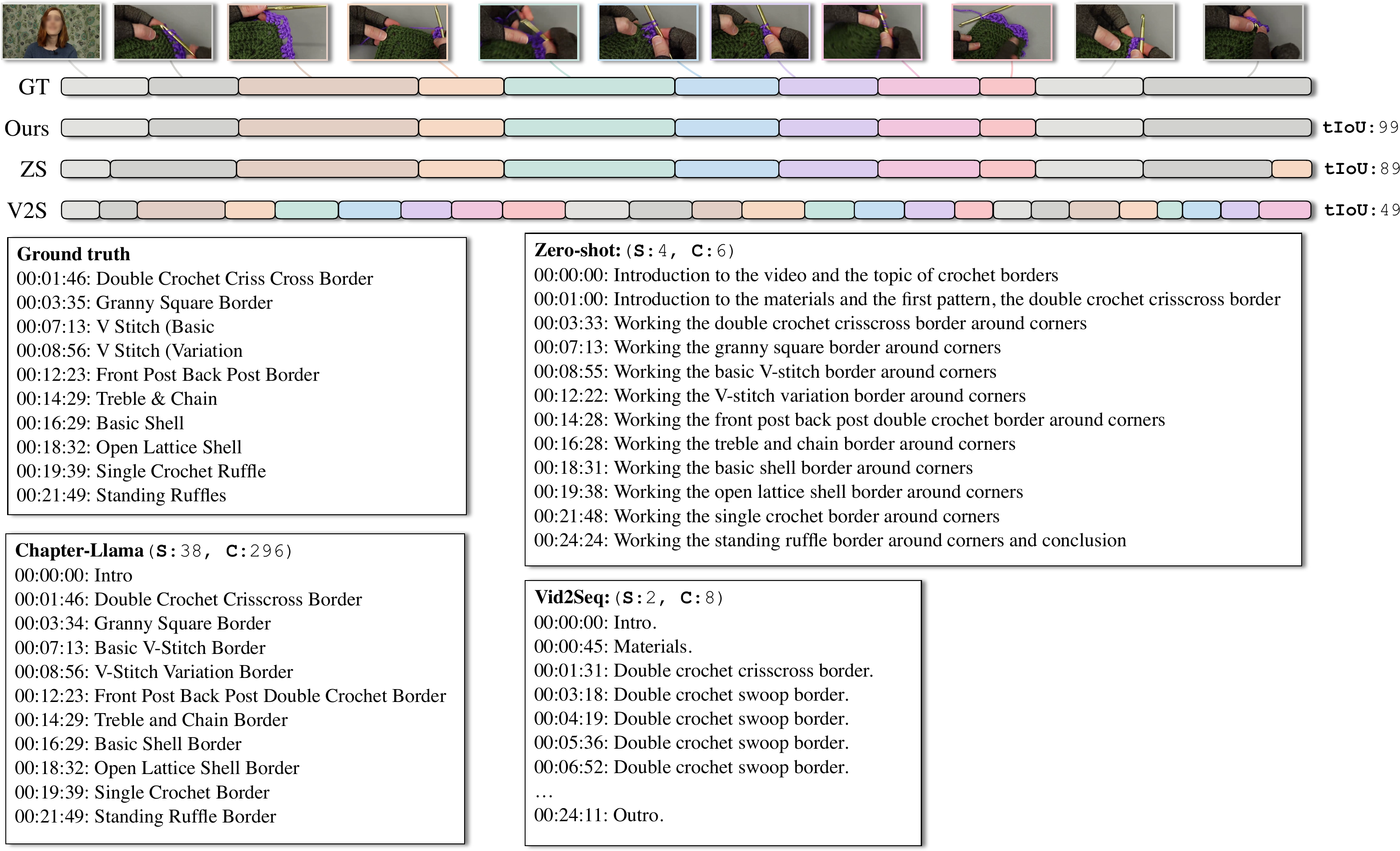}
	\includegraphics[width=.99\linewidth]{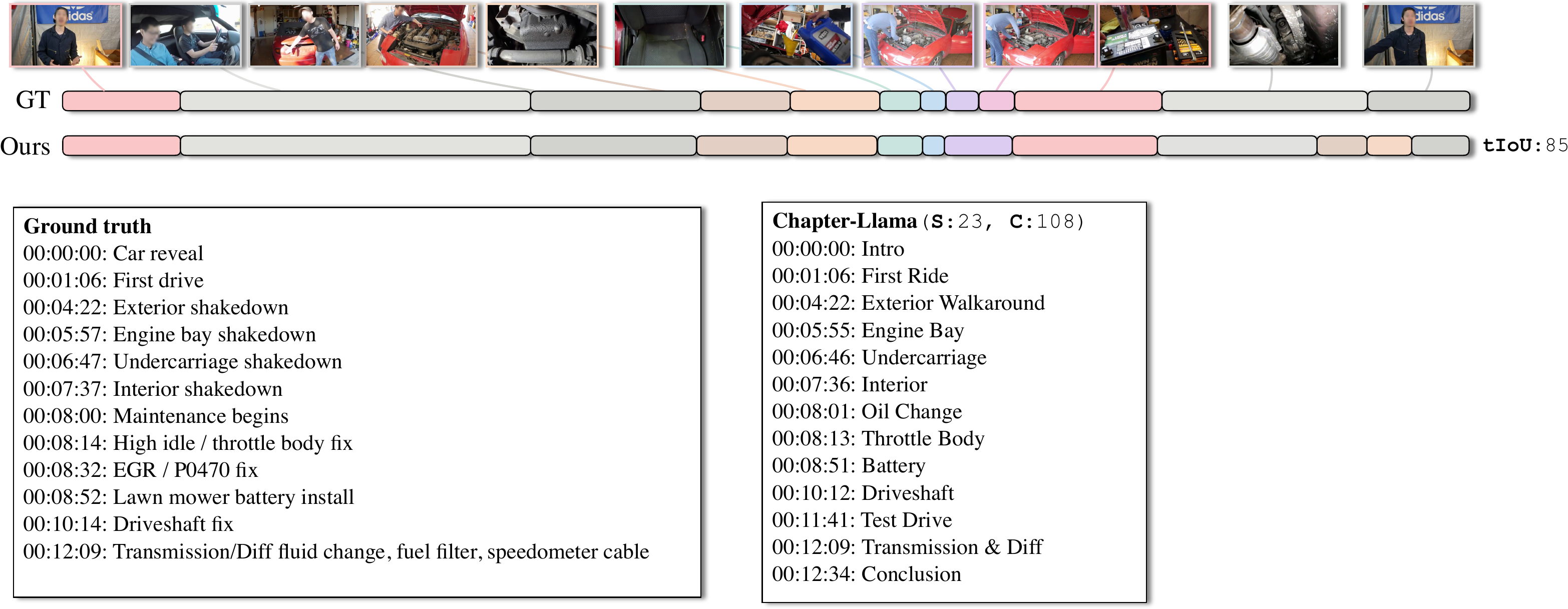}
	\caption{\textbf{Additional qualitative examples:} 
	We show two more examples of our \method predictions compared to the ground truth (GT). 
	Our method generates accurate temporal boundaries and relevant chapter titles that align well with the video content.
    For each example, we display the corresponding SODA (S) and CIDEr (C) scores.
	}
	\label{fig:app:qualitative}
\end{figure*}

Similar to \appendixref{Fig.~3}{\cref{fig:qualitative}} of the main paper,
in \cref{fig:app:qualitative},
we present two additional examples comparing our method against Vid2Seq and our zero-shot baseline.

\begin{figure*}
	\centering
	\includegraphics[width=.99\linewidth]{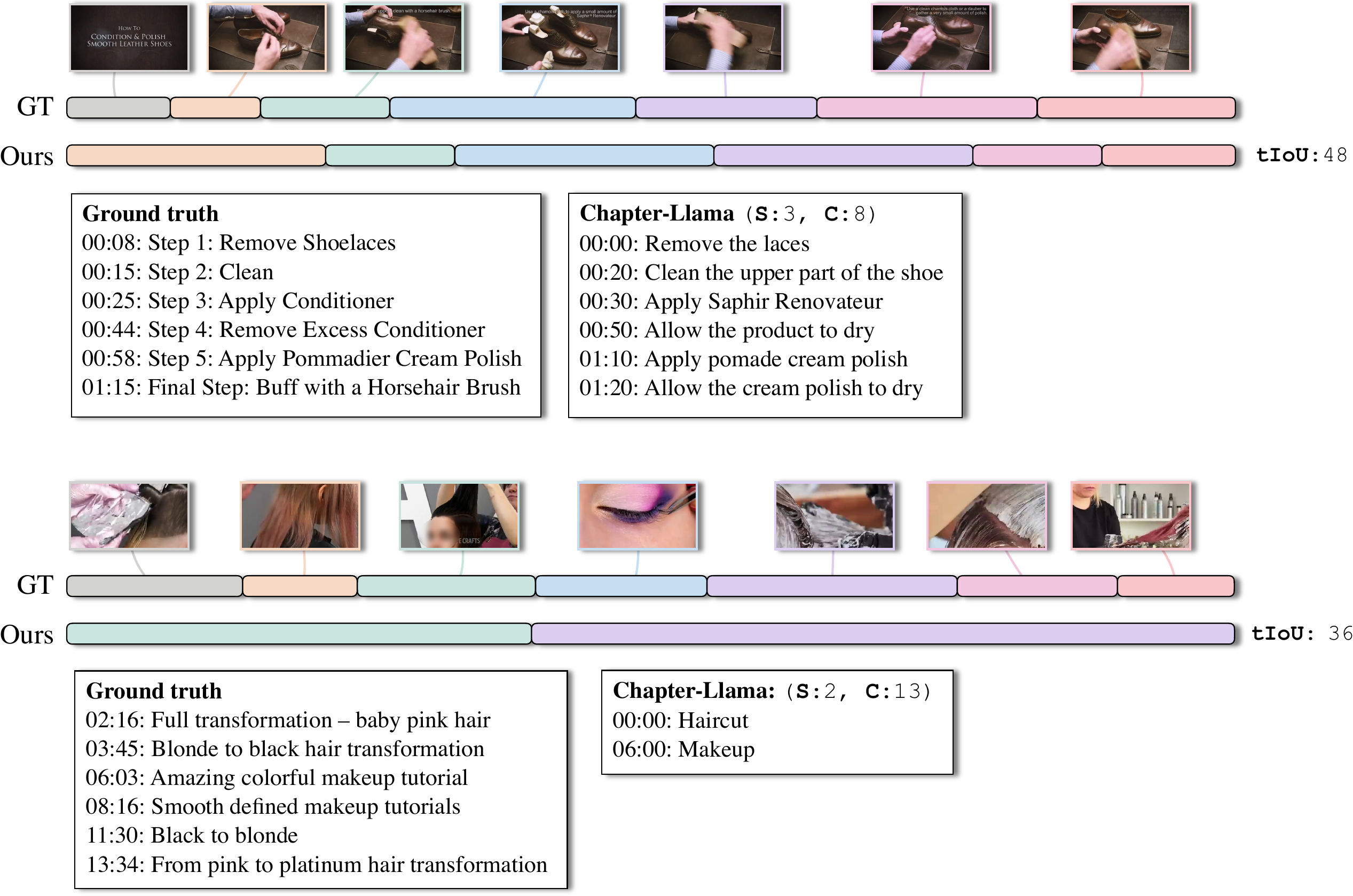}
	
	\vspace{0.6cm}

	\includegraphics[width=.99\linewidth]{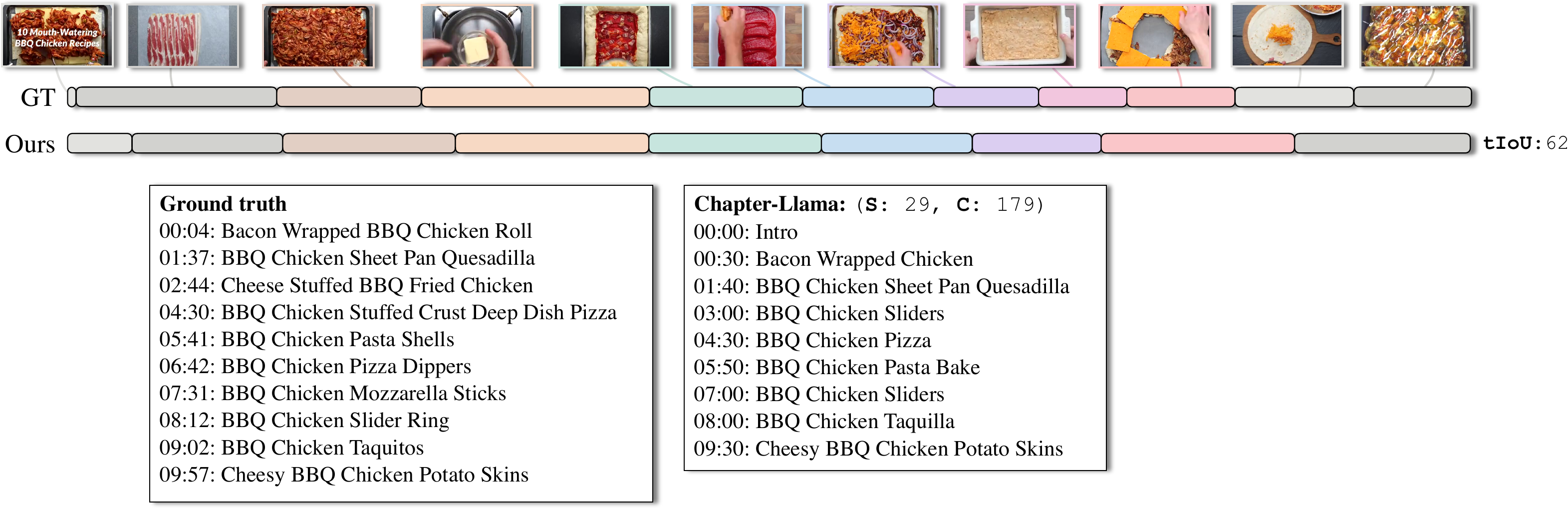}
	\caption{\textbf{\new{Additional qualitative examples without ASR:}}
	\new{We show three examples of videos without speech, 
	comparing our \method predictions to ground truth (GT). 
	Despite lacking ASR,
	our method still produces reasonable chapters 
	by leveraging visual cues and on-screen text when available
	(top and bottom examples).
    For each example, we display the corresponding SODA (S) and CIDEr (C) scores.}
	}
	\label{fig:app:qualitative_no-asr}
\end{figure*}

\new{In \cref{fig:app:qualitative_no-asr}, 
we show three examples of our \method predictions compared to the ground truth (GT) 
for videos without speech (3\% of the data).
We observe that many of the completely `speechless' videos contain OCR-readable text
to help the viewer follow the video (top and bottom examples), 
in which cases the captioners tend to perform OCR, leading to satisfactory chaptering results.
Otherwise, in case of no on-screen text and no speech (e.g., only music), the result is inferior, 
though still acceptable (middle example).
As also evaluated in \cref{tab:app:no_asr_val}, 
our model still achieves reasonable quantitative performance, 
even if speech indeed tends to be more informative for chaptering than visual modality~\cite{yang2023vidchapters}.}
	
\end{document}